\documentclass[10pt,twocolumn,letterpaper]{article}
\usepackage[pagenumbers]{cvpr} %

\usepackage{amsmath,amsfonts,bm}

\def\eqref#1{equation~\ref{#1}}

\def\1{\bm{1}}

\def\vzero{{\bm{0}}}
\def\vone{{\bm{1}}}

\def\vc{{\bm{c}}}

\def\vt{{\bm{t}}}
\def\vu{{\bm{u}}}
\def\vv{{\bm{v}}}

\def\vx{{\bm{x}}}

\def\vz{{\bm{z}}}

\def\mI{{\bm{I}}}

\DeclareMathAlphabet{\mathsfit}{\encodingdefault}{\sfdefault}{m}{sl}
\SetMathAlphabet{\mathsfit}{bold}{\encodingdefault}{\sfdefault}{bx}{n}

\newcommand{\vepsilon}{\bm{\epsilon}}

\usepackage{pifont}
\usepackage{xspace}
\usepackage{tikz}
\usepackage{algorithm}
\usepackage{algpseudocode}
\definecolor{cgreen}{HTML}{1E8E3E}
\definecolor{cred}{HTML}{D93025}\newcommand{\cmark}{\textcolor{cgreen}{\ding{51}}}
\newcommand{\xmark}{\textcolor{cred}{\ding{55}}}
\newcommand{\method}{CSF\xspace}

\definecolor{cvprblue}{rgb}{0.21,0.49,0.74}
\usepackage[pagebackref,breaklinks,colorlinks,allcolors=cvprblue]{hyperref}
\def\confName{CVPR}
\def\confYear{2027}
\title{Causal Self Forcing: Causal Teachers Beat Bidirectional Teachers \\ in Interactive Video Model Distillation}
\author{
    Jingyang Ou$^{1,2,3,4,*}$\quad
    Ziheng Chen$^{4}$\quad
    Hongzhou Zhu$^{5}$\quad
    Liangyu Yuan$^{4,6}$\quad
    Yicheng Ji$^{4,7}$\\
    Wenbo Su$^{4}$\quad
    Bo Zheng$^{4}$\quad
    Chongxuan Li$^{1,2,3,\dagger}$\\[4pt]
    {\normalsize $^{1}$Gaoling School of Artificial Intelligence, Renmin University of China}\\
    {\normalsize $^{2}$Beijing Key Laboratory of Research on Large Models and
    Intelligent Governance}\\
    {\normalsize $^{3}$Engineering Research Center of Next-Generation Intelligent
    Search and Recommendation, MOE}\\
    {\normalsize $^{4}$Alibaba Group\quad
    $^{5}$Tsinghua University\quad
    $^{6}$Shanghai Jiao Tong University\quad
    $^{7}$Zhejiang University}
}
\begin{document}
\maketitle
{\renewcommand{\thefootnote}{\fnsymbol{footnote}}%
    \footnotetext[1]{Work done during Jingyang Ou's internship at Alibaba Group.}%
    \footnotetext[2]{Corresponding author: \href{mailto:chongxuanli@ruc.edu.cn}{\texttt{chongxuanli@ruc.edu.cn}}.}}
\begin{abstract}
    Current few-step autoregressive video world models are predominantly
    distilled with Distribution Matching Distillation (DMD) from bidirectional
    teachers, which are architecturally mismatched with their causal students.
    Causal teachers avoid this mismatch, yet they have so far yielded weaker
    students. This gap is commonly attributed to the teacher's own poor
    sampling: errors accumulate over multi-step autoregressive generation and
    are thought to propagate from the teacher to the student.
    We challenge this belief with a simple observation: in DMD, the causal teacher
    never generates videos itself; it only scores student rollouts.
    Whereas sampling on clean context accumulates errors, scoring does not have to:
    such error propagation can be substantially mitigated if the teacher avoids
    scoring student rollouts under a clean yet drifted context.
    Motivated by this insight, we propose Causal Self Forcing (CSF), which lets a
    Diffusion-Forcing (DF) teacher score on-policy student rollouts under a
    heavily noised history context, where rollout errors are largely masked
    rather than propagated.
    Built on Wan2.2-5B, our few-step CSF student outperforms strictly matched
    distillation baselines with either bidirectional or causal teachers, and
    surpasses its own multi-step teacher.
    In minute-long generation, its advantage widens further, and it even matches or
    exceeds substantially larger world models.
\end{abstract}

\section{Introduction}
\label{sec:intro}

Large-scale video diffusion
models~\cite{peebles2023scalable,yang2024cogvideox,polyak2024movie,kong2024hunyuanvideo,hacohen2024ltx,wan2025}
have learned rich visual and physical priors, driving the shift from offline
video synthesis to real-time interactive video world
models~\cite{bruce2024genie,alonso2024diamond,valevski2024gamengen,sun2025worldplay,gao2026lingbot,zhao2026minwm,huang2026solarwm}.
Given an initial visual observation, an interactive world model generates future
frames online in response to streaming user
actions~\cite{guo2025mineworld,zhang2025matrixgame,he2025matrixgame2,qian2026matrixgame35,team2026dreamx}
or camera
trajectories~\cite{wang2024motionctrl,he2024cameractrl,ren2025gen3c,li2026prope,zhang2026worldplay2}.
Unlike offline video diffusion, real-time
interactive simulation imposes two requirements:
(i)~\emph{block-causal generation}~\cite{chen2024diffusion,jin2024pyramidal,gao2024ca2,song2025history,teng2025magi},
where each new frame block is synthesized sequentially from cached history and
current controls; and
(ii)~\emph{few-step
    inference}~\cite{song2023consistency,yin2024one,yin2024improved,lu2024simplifying,gu2026anyflow},
where multi-step denoising is compressed into a few network evaluations to meet
strict latency budgets.

To meet both requirements, Distribution Matching Distillation
(DMD)~\cite{yin2024one,yin2024improved}, first applied to autoregressive video
models by CausVid~\cite{yin2025causvid} and made on-policy by Self
Forcing (SF)~\cite{huang2025selfforcing} to mitigate exposure bias~\cite{guo2025end},
has become the standard post-training paradigm for interactive world
models~\cite{yang2025longlive,liu2025rolling,cui2025self,lin2026autoregressive,feng2026one,chen2026contextforcing,zhuang2026self}.
Yet these methods still score the causal student's rollouts with a
full-sequence \emph{bidirectional teacher}, which is architecturally
mismatched with the student.
This mismatch incurs two costs.
First, because a block-causal student cannot be directly initialized from a
bidirectional teacher~\cite{zhu2026causalforcing}, existing
pipelines~\cite{sun2025worldplay,gao2026lingbot,zhu2026sanawm,zhao2026minwm,gao2026lingbot2,huang2026solarwm}
require three to four stages with intermediate causal
diffusion~\cite{teng2025magi,gu2026anyflow,huang2026solarwm} and few-step
initialization~\cite{zhu2026causalforcing,zhao2026causalforcingpp,zheng2026causalrcm,zhao2026minwm}
(\cref{fig:two_stage_pipeline}, top).
Second, bidirectional scoring conditions each block on future frames and
controls that the causal student never sees.
A natural remedy is \emph{Causal DMD}, which distills with a block-causal
teacher that matches the student and thus avoids both costs.
Yet causal teachers have so far yielded weaker students~\cite{yin2025causvid}.
This gap is commonly attributed to the teacher's own poor sampling: errors
accumulate over multi-step autoregressive generation and are thought to
propagate from the teacher to the student.

\begin{table}[t]
    \centering
    \footnotesize
    \setlength{\tabcolsep}{0pt}
    \caption{\textbf{Comparison of key properties across autoregressive video
            distillation methods.} Reliable teacher scoring: the teacher
        scores student rollouts in a regime where its score is well trained.
        N/A for CausVid, which conditions the student on noised ground-truth
        context rather than its own rollouts.}
    \label{tab:gradient_comparison}
    \begin{tabular*}{\linewidth}{@{\extracolsep{\fill}}lcccc@{}}
        \toprule
        & CausVid & Self Forcing & Full CMD & \textbf{\method} \\[-1pt]
        Property
        & \cite{yin2025causvid}
        & \cite{huang2025selfforcing}
        & \cite{bandyopadhyay2026cmd}
        & \textbf{(Ours)} \\
        \midrule
        On-policy rollout & \xmark & \cmark & \cmark & \cmark \\
        Causal teacher & \xmark & \xmark & \cmark & \cmark \\
        Reliable teacher scoring & \textcolor{gray}{N/A} & \cmark & \xmark & \cmark \\
        \midrule
        Scoring token length
        & \textcolor{cgreen}{\textbf{$1\times$}}
        & \textcolor{cgreen}{\textbf{$1\times$}}
        & \textcolor{cred}{\textbf{$2\times$}}
        & \textcolor{cgreen}{\textbf{$1\times$}} \\
        \bottomrule
    \end{tabular*}
\end{table}

We challenge this belief with a simple observation: \emph{in DMD, the teacher
    never generates videos itself; it only scores student rollouts}.
Whereas sampling conditions on clean yet imperfect frames and accumulates
errors (\cref{fig:teacher_student_qualitative}, top), scoring does not have
to: such error propagation can be substantially mitigated if the teacher
avoids scoring student rollouts under a clean yet drifted context.
Such scoring is unavoidable for a Teacher-Forcing (TF) teacher, which is
trained solely on clean context.
A Diffusion-Forcing (DF) teacher~\cite{chen2024diffusion} learns scores
across context noise levels, yet prior Causal DMD, represented by the recent
Full CMD~\cite{bandyopadhyay2026cmd}, still scores under clean or low-noise
context.
Under clean-context scoring, the student merely tracks its teacher, whether TF
or DF (\cref{fig:sec3_wbench}).
Querying the DF teacher under \emph{high context noise} instead largely masks rollout
errors and evaluates the teacher where it is well trained, thereby
\emph{decoupling distillation from sampling}.

Motivated by this insight, we propose \emph{Causal Self Forcing (\method)},
which lets a DF teacher score on-policy student
rollouts~\cite{huang2025selfforcing} under heavily noised context
(\cref{tab:gradient_comparison}).
\method perturbs each rollout with the high-noise-shifted schedule of DF
pretraining into a \emph{single} noisy sequence, where
each block serves as both its own scoring target and part of the noisy context
for later blocks in one block-causal pass (\cref{fig:df_dmd_comparison}).
This yields three benefits:
(i)~\emph{reliable teacher scoring}, with rollout errors largely masked and the
teacher queried where it is well trained;
(ii)~\emph{causal supervision} free of future frames and controls; and
(iii)~\emph{a two-stage world model pipeline} (\cref{fig:two_stage_pipeline},
bottom), where Stage~1 trains control-conditioned causal DF and Stage~2
directly initializes the teacher, critic, and student from that single
checkpoint.

Extensive experiments validate the effectiveness of \method.
Under strictly matched settings on Wan2.2-5B~\cite{wan2025}, our few-step
student achieves $75.7$ on WBench~\cite{ying2026wbench}, outperforming both
the bidirectional-teacher SolarWM-5B~\cite{huang2026solarwm} ($72.8$) and the
causal-teacher Full CMD~\cite{bandyopadhyay2026cmd} ($72.3$), and also
surpassing its own multi-step teacher ($69.6$).
In 60\,s generation on the SANA-WM benchmark~\cite{zhu2026sanawm}, its
advantage widens further, and it even matches or exceeds the substantially
larger HY-WorldPlay (8B)~\cite{sun2025worldplay} and LingBot-World-Fast
(14B)~\cite{gao2026lingbot}, attaining the highest VBench score among all
compared methods.

In summary, our contributions are as follows:
\begin{itemize}
    \item We challenge the prevailing belief that causal teachers inherently
          yield weaker students in DMD. Since the teacher only scores student
          rollouts, the propagation of its sampling errors can be
          substantially mitigated if it avoids scoring under a clean yet
          drifted context, as our analysis of TF and DF teachers confirms
          (\cref{fig:sec3_wbench}).
    \item We propose \method, which scores on-policy student rollouts with a
          DF teacher under high context noise in a single block-causal pass.
          It provides fully causal supervision and reduces the world model
          pipeline to two stages.
    \item We demonstrate that, under strictly matched settings, \method
          outperforms distillation with either bidirectional or causal teachers
          on WBench and in our user study, surpasses its own multi-step
          teacher, and matches or exceeds substantially larger models in
          minute-long generation.
\end{itemize}

\section{Background}
\label{sec:background}

\subsection{Diffusion Models}
\label{sec:background_diffusion}
Diffusion models~\cite{sohldickstein2015deep,ho2020denoising} learn to transport samples between a data distribution and a simple noise distribution.
Let $\vx_0 \sim p_{\mathrm{data}}$ be a data sample,
$\vepsilon \sim \mathcal{N}(\vzero,\mI)$ be standard Gaussian noise,
and $t\in[0,1]$ be a diffusion timestep.
Under the standard linear schedule in flow matching~\cite{lipman2023flow}, the noisy interpolant at timestep $t$ is
$\vx_t = (1-t)\vx_0 + t\vepsilon$,
and a velocity network $v_\theta(\vx_t,t,\vc)$ is trained
to predict the target velocity $\vu_t=\vepsilon-\vx_0$
given optional conditioning $\vc$:
\begin{equation}
    \mathcal{L}_{\mathrm{FM}}(\theta)
    = \mathbb{E}_{\vx_0,\vepsilon,t}\!\Big[
        \lambda(t)\big\|v_\theta(\vx_t,t,\vc)-\vu_t\big\|_2^2\Big].
    \label{eq:flow_matching}
\end{equation}

With the learned velocity network, a sample $\vx_0$ can be generated by solving the following ODE from $t=1$ to $0$:
\begin{equation}
    \frac{d\vx_t}{dt} = v_\theta(\vx_t,t,\vc), \quad \vx_1\sim\mathcal{N}(\vzero,\mI).
    \label{eq:diffusion_ode}
\end{equation}
Alternatively, the velocity network can be reparameterized as a score network~\cite{song2021score}
$s_\theta(\vx_t,t,\vc) = \nabla_{\vx_t}\log p_{\theta}(\vx_t\mid t, \vc)$:
\begin{equation}
    s_\theta(\vx_t,t,\vc)
    = -\frac{\vx_t+(1-t)v_\theta(\vx_t,t,\vc)}{t}.
    \label{eq:velocity_to_score}
\end{equation}

\subsection{Autoregressive Video Diffusion Models}
\label{sec:background_ar}
We partition a video latent sequence into $N$ temporal blocks,
$\vx^{1:N}=(\vx^1,\ldots,\vx^N)$, where each block contains one or more frames.
Given blockwise controls $\vc^{1:N}$ and an optional clean initial frame
$\vx^0$\footnote{Throughout training and distillation, $\vx^0$ remains clean
    without added noise; we omit $\vx^0$ in notation for brevity.},
a block-causal autoregressive (AR) video model factorizes the joint distribution
as
\begin{equation}
    p_\theta(\vx^{1:N}\mid\vc^{1:N})
    =\prod_{i=1}^{N}p_\theta(\vx^i\mid\vx^{<i},\vc^{\leq i}),
    \label{eq:ar_factorization}
\end{equation}
where each conditional distribution $p_\theta(\vx^i\mid\vx^{<i},\vc^{\leq i})$ is
modeled by a diffusion model.
Existing methods typically learn this conditional distribution via one of the following
two training strategies:

\paragraph{Teacher Forcing (TF).}
Teacher forcing~\cite{williams1989learning} trains the model to denoise each target block $\vx^i_{t_i}$
conditioned on the clean ground-truth (GT) context $\vx^{<i}$ at zero noise:
\begin{equation}
    \thinmuskip=2mu \medmuskip=2mu \thickmuskip=3mu
    \mathcal{L}_{\text{\tiny TF}}(\theta)
    =\mathbb{E}\!\Big[\sum_{i=1}^{N}\!\lambda(t_i)
    \big\|v_\theta(\vx^i_{t_i},t_i\mid\vx^{<i},\vzero,\vc^{\leq i})-\vu^i_{t_i}\big\|_2^2\Big].
    \label{eq:teacher_forcing}
\end{equation}
As each block serves both as a noisy denoising target and as
clean context for subsequent blocks, parallel training under a
block-causal attention mask concatenates the clean sequence with its noisy
counterpart, doubling the sequence length ($2\times$ tokens).

\paragraph{Diffusion Forcing (DF).}
Diffusion Forcing~\cite{chen2024diffusion} assigns independent random
timesteps $\vt=(t_1,\ldots,t_N)$ and Gaussian noises
$\vepsilon^i\sim\mathcal{N}(\vzero,\mI)$ across blocks:
\begin{equation}
    \vx^i_{t_i}=(1-t_i)\vx^i+t_i\vepsilon^i.
    \label{eq:df_corruption}
\end{equation}
In practice, DF training~\cite{zheng2026causalrcm,teng2025magi} draws each
$t_i$ from a high-noise-shifted distribution.\footnote{Timesteps are sampled via
    $t = \frac{s u}{1+(s-1)u}$ with $u\sim\mathcal{U}[0,1]$ and shift $s>1$,
    e.g., $s=3$ in~\cite{teng2025magi} and $s=5$ in~\cite{zheng2026causalrcm}.}
Under a block-causal attention mask, the network predicts the velocity of
each block $i$ conditioned on its randomly corrupted causal context
$\vx_{\vt}^{<i}$ within the same sequence:
\begin{equation}
    \thinmuskip=2mu \medmuskip=2mu \thickmuskip=3mu
    \mathcal{L}_{\text{\tiny DF}}(\theta)
    =\mathbb{E}\!\Big[\sum_{i=1}^{N}\!\lambda(t_i)
    \big\|v_\theta(\vx^i_{t_i},t_i\mid\vx_{\vt}^{<i},\vt^{<i},\vc^{\leq i})-\vu^i_{t_i}\big\|_2^2\Big].
    \label{eq:diffusion_forcing}
\end{equation}
At inference time, both TF- and DF-trained models sample sequentially following
\cref{eq:ar_factorization}, solving \cref{eq:diffusion_ode} for each block
$\hat{\vx}^i$ conditioned on previously generated \emph{clean} blocks
($\vt^{<i}=\vzero$) to preserve frame-to-frame continuity.
See Appendix~\ref{app:related_work} for additional related work.

\subsection{Distribution Matching Distillation}
\label{sec:background_dmd}
Distribution Matching Distillation (DMD)~\cite{wang2023prolificdreamer,yin2024one,yin2024improved}
distills a pretrained diffusion teacher $s_{\mathrm{real}}$ into a few-step
generator $G_\theta$ by minimizing the reverse KL divergence
$D_{\mathrm{KL}}(p_{\theta,t}\Vert p_{\mathrm{real},t})$ across timesteps.
Given a student sample $\hat{\vx}=G_\theta(\vz,\vc)$ and its noisy interpolant
$\hat{\vx}_t=(1-t)\hat{\vx}+t\vepsilon$, let
$\Delta s(\hat{\vx}_t,t,\vc)\mathrel{:=}
    s_{\mathrm{fake}}(\hat{\vx}_t,t,\vc)-s_{\mathrm{real}}(\hat{\vx}_t,t,\vc)$
denote the score difference between a trainable critic $s_{\mathrm{fake}}$ and
the frozen teacher $s_{\mathrm{real}}$.
The generator is updated via
\begin{equation}
    \nabla_\theta\mathcal{L}_{\mathrm{DMD}}
    =\mathbb{E}\!\left[
        w(t)\left(\frac{\partial\hat{\vx}}{\partial\theta}\right)^{\!\top}\!\Delta s(\hat{\vx}_t,t,\vc)
        \right],
    \label{eq:dmd_gradient}
\end{equation}
while $s_{\mathrm{fake}}$ is
trained in alternating critic updates on detached student samples $\hat{\vx}$ using
\cref{eq:flow_matching}.

\section{Rethinking DMD with Causal Teachers}
\label{sec:rethink}

\newcommand{\wbTFteacher}{69.4} %
\newcommand{\wbDFteacher}{69.6} %
\newcommand{\wbBi}{72.8}        %
\newcommand{\wbTFclean}{70.3}   %
\newcommand{\wbDFclean}{68.2}   %
\newcommand{\wbDFours}{75.7}    %
\definecolor{wbDF}{HTML}{E07B24} %
\definecolor{wbTF}{HTML}{2A9D8F} %
\definecolor{wbBiC}{HTML}{5B6B82} %

\begin{figure}[t]
    \centering
    \begin{tikzpicture}[
        x=1cm, y=1cm, font=\scriptsize,
        every node/.style={inner sep=0pt},
        head/.style={anchor=base west, font=\scriptsize\bfseries, text=black!70},
        lab/.style={anchor=base east, yshift=-2.3pt},
        val/.style={anchor=base west, yshift=-2.3pt},
    ]
        \def\wbxmin{55}     %
        \def\wbu{0.19}      %
        \def\wbx#1{(#1-\wbxmin)*\wbu}
        \def\wbW{4.50}      %
        \def\wbxL{-3.30}    %
        \def\wbh{0.13}      %
        \def\wbybot{0.58}\def\wbytop{4.04}

        \def\wbbar#1#2#3#4#5{%
            \fill[#3, rounded corners=1.3pt] (0,{#1-\wbh}) rectangle ({\wbx{#2}},{#1+\wbh});
            \fill[#3] (0,{#1-\wbh}) rectangle (0.08,{#1+\wbh});
            \node[lab, font=#5] at (-0.10,#1) {#4};
            \node[val, font=#5] at ({\wbx{#2}+0.07},#1) {#2};
        }
        \def\wbbartwo#1#2#3#4#5{%
            \fill[#3, rounded corners=1.3pt] (0,{#1-\wbh}) rectangle ({\wbx{#2}},{#1+\wbh});
            \fill[#3] (0,{#1-\wbh}) rectangle (0.08,{#1+\wbh});
            \node[lab] at (-0.10,{#1+0.12}) {#4};
            \node[lab, text=black!65] at (-0.10,{#1-0.12}) {#5};
            \node[val] at ({\wbx{#2}+0.07},#1) {#2};
        }

        \node[head] at (\wbxL,4.14) {Multi-step Causal Teachers};
        \wbbar{3.78}{\wbTFteacher}{wbTF!30}{\textcolor{wbTF!80!black}{TF Teacher}}{\scriptsize}
        \wbbar{3.38}{\wbDFteacher}{wbDF!28}{\textcolor{wbDF!80!black}{DF Teacher}}{\scriptsize}

        \node[head] at (\wbxL,2.84) {Few-step Distilled Students};
        \wbbar{2.48}{\wbBi}{wbBiC!75}{\textcolor{wbBiC}{Bi-Teacher} + SF}{\scriptsize}
        \wbbartwo{1.97}{\wbTFclean}{wbTF!80}{\textcolor{wbTF!80!black}{TF Teacher} + TF-style DMD}{(clean context)}
        \wbbartwo{1.41}{\wbDFclean}{wbDF!58}{\textcolor{wbDF!80!black}{DF Teacher} + TF-style DMD}{(clean context)}
        \wbbar{0.90}{\wbDFours}{wbDF}{\textcolor{wbDF!80!black}{DF Teacher} + \method (Ours)}{\scriptsize\bfseries}

        \draw[black!60, line width=0.4pt] (0,\wbytop) -- (0,\wbybot) -- (\wbW,\wbybot);
        \foreach \s in {55, 60, 65, 70, 75} {
            \draw[black!60, line width=0.4pt] ({\wbx{\s}},\wbybot) -- ({\wbx{\s}},{\wbybot-0.05});
            \node[anchor=north, text=black!70] at ({\wbx{\s}},{\wbybot-0.10}) {\s};
        }
        \node[anchor=north] at ({0.5*\wbW},{\wbybot-0.42}) {WBench Overall Score ($\uparrow$)};
    \end{tikzpicture}

    \caption{\textbf{WBench~\cite{ying2026wbench} overall scores of multi-step
    causal teachers and few-step distilled students.}
    ``Bi-Teacher + SF'' denotes the released SolarWM-5B
    checkpoint~\cite{huang2026solarwm}, which is distilled with Self Forcing
    (SF) under a bidirectional teacher.
    TF-style DMD scores each noisy block against a separate context sequence,
    which is kept clean ($\vt^{<i}=\vzero$) here.}
    \label{fig:sec3_wbench}
\end{figure}
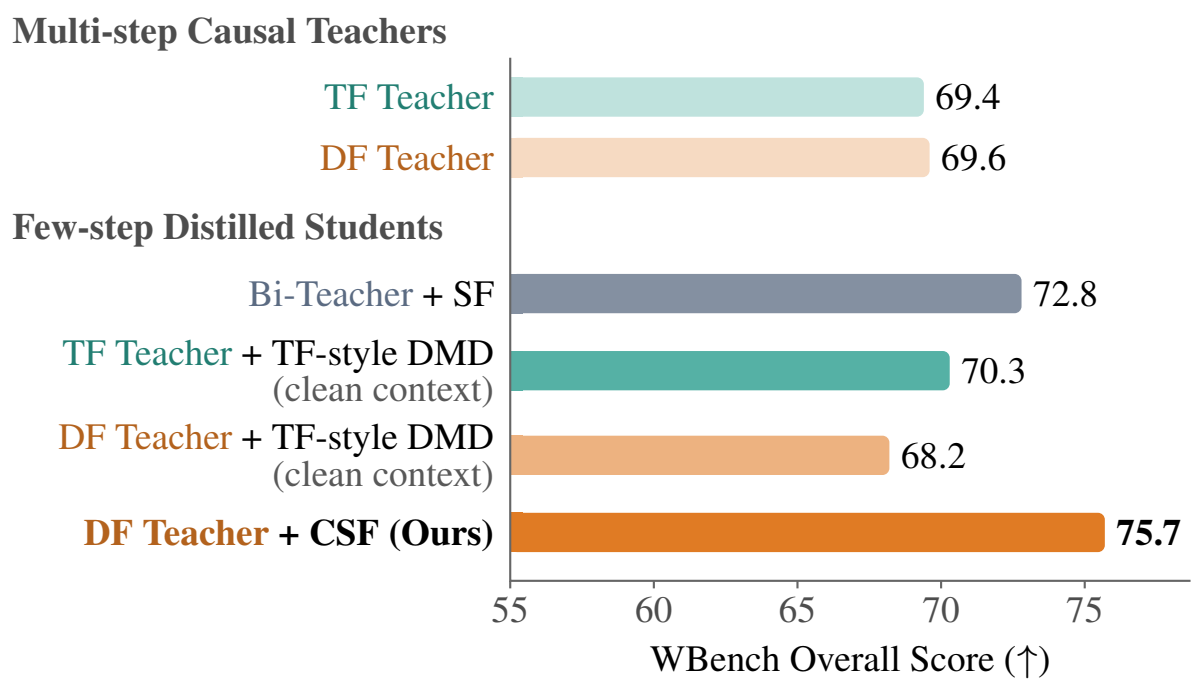

\begin{figure}[t]
    \centering
    \setlength{\tabcolsep}{1pt}
    \renewcommand{\arraystretch}{0.9}
    \begin{tabular}{@{}c c c c@{}}
        & \scriptsize 1\,s & \scriptsize 5\,s & \scriptsize 11\,s \\
        \raisebox{1.7ex}{\shortstack{\scriptsize DF\\\scriptsize teacher}} &
        \includegraphics[width=0.285\linewidth]{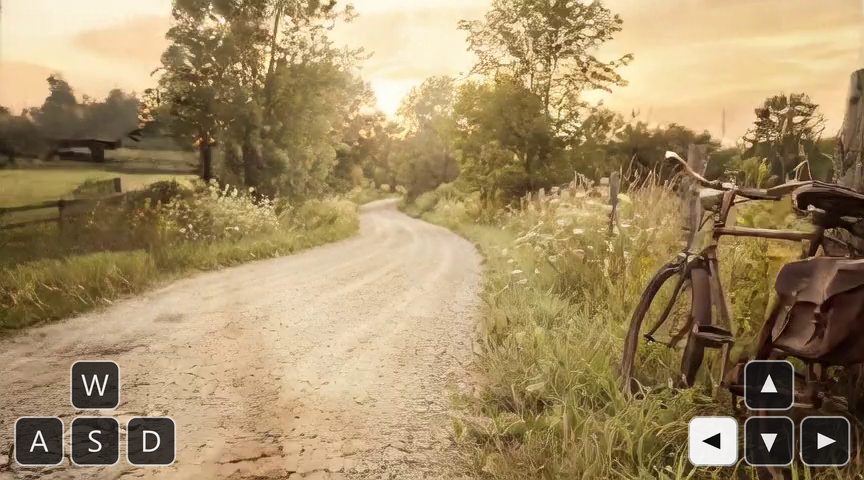} &
        \includegraphics[width=0.285\linewidth]{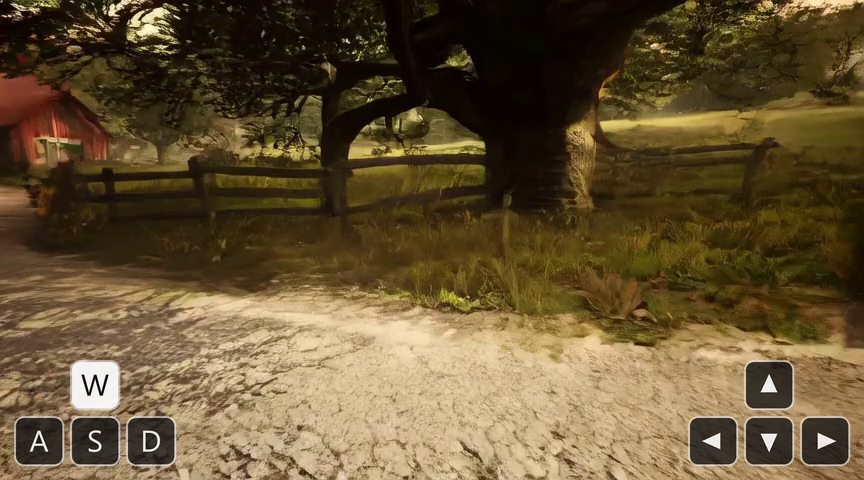} &
        \includegraphics[width=0.285\linewidth]{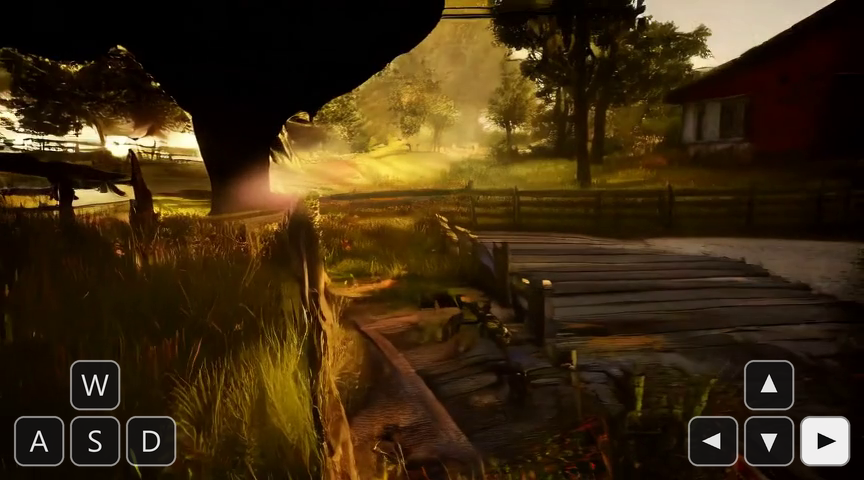} \\
        \raisebox{1.7ex}{\shortstack{\scriptsize \method\\\scriptsize student}} &
        \includegraphics[width=0.285\linewidth]{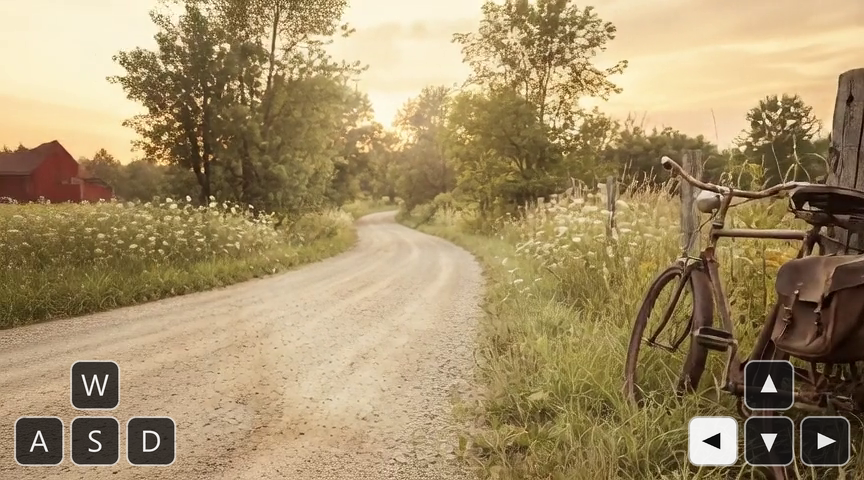} &
        \includegraphics[width=0.285\linewidth]{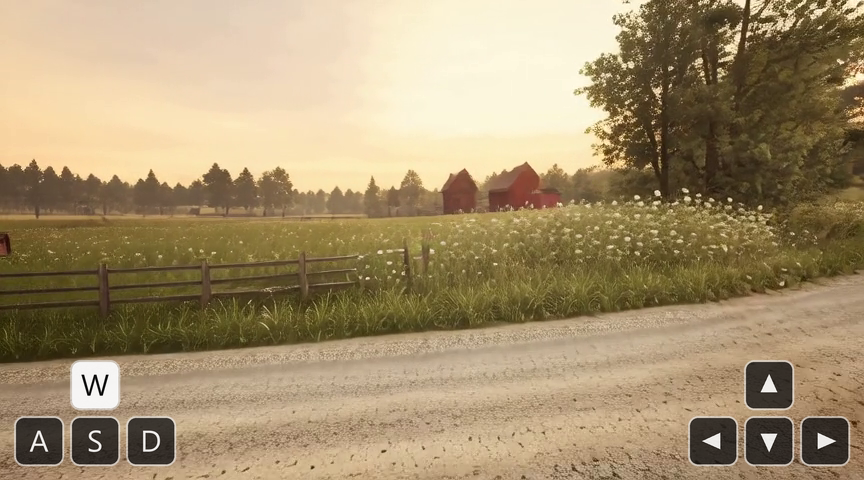} &
        \includegraphics[width=0.285\linewidth]{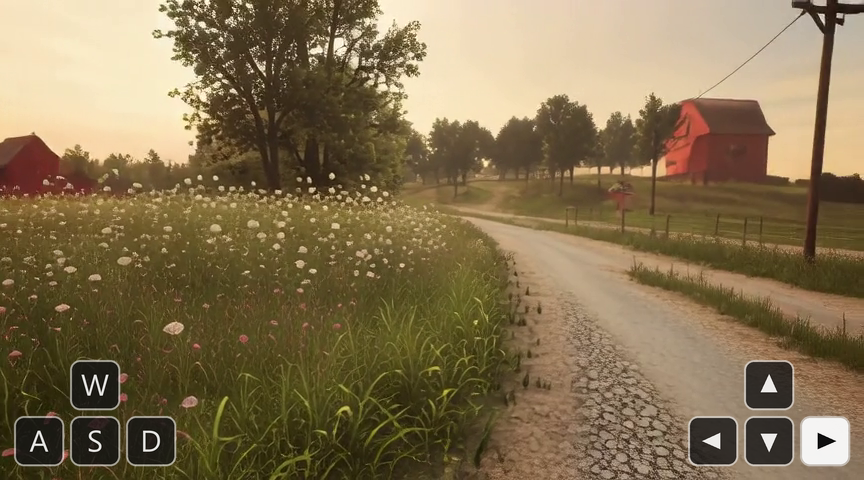}
    \end{tabular}
    \caption{\textbf{Rollouts of a DF teacher and its \method student} from
        the same input image and camera trajectory.
        \textbf{Top:} multi-step sampling of the DF teacher accumulates errors
        and develops structural distortions, especially beyond the 5\,s
        training length.
        \textbf{Bottom:} the few-step student distilled from the same teacher
        does not inherit this drift and preserves scene details.}
    \label{fig:teacher_student_qualitative}
\end{figure}

\begin{figure*}[t]
    \centering
    \includegraphics[width=\textwidth]{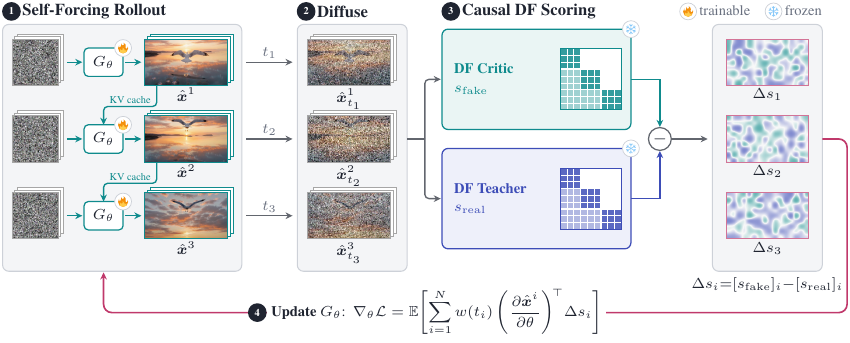}
    \caption{\textbf{Overview of Causal Self Forcing (\method)}.
        \textbf{(1)}~The causal student $G_\theta$ generates an on-policy rollout
        $\hat{\vx}^{1:N}$ on its own clean history.
        \textbf{(2)}~Each block is diffused with high noise $t_i$ into a single
        sequence $\hat{\vx}_{\vt}^{1:N}$ ($1\times$ tokens).
        \textbf{(3)}~The DF teacher $s_{\mathrm{real}}$ and DF critic
        $s_{\mathrm{fake}}$ score it in one block-causal forward pass (dark: scored
        block; light: noisy context).
        \textbf{(4)}~The per-block difference $\Delta s_i$ updates $G_\theta$.
        Only the generator update is shown; $s_{\mathrm{fake}}$ is trained in
        alternating critic updates. Controls $\vc^{1:N}$ are omitted for clarity.}
    \label{fig:df_dmd_comparison}
\end{figure*}
As discussed in \cref{sec:intro}, mainstream
methods~\cite{huang2025selfforcing,zhu2026causalforcing,zhuang2026self}
distill the causal student from a bidirectional teacher, a setting we term
\emph{asymmetric DMD}.
Its causal counterpart, Causal DMD, avoids the architectural mismatch with a
block-causal teacher, yet is commonly believed to be
inferior~\cite{yin2025causvid}: a causal teacher accumulates errors over
multi-step autoregressive sampling, and these errors are thought to propagate
to its distilled student.

At first glance, this belief seems natural: if distillation imitated teacher
trajectories~\cite{song2023consistency,lu2024simplifying}, the teacher's
rollout drift would directly limit the student.
Yet it overlooks a key fact: \emph{in DMD, the teacher never generates videos
    itself; it only scores student rollouts}.
The student never observes teacher-generated frames; across $N$ temporal
blocks, $G_\theta$ is updated solely through the per-block score differences
$\Delta s_i=[s_{\mathrm{fake}}]_i-[s_{\mathrm{real}}]_i$ evaluated on its own
rollouts, where $[\cdot]_i$ denotes the score of block~$i$:
\begin{equation}
    \nabla_\theta\mathcal{L}
    =
    \mathbb{E}\!\left[
        \sum_{i=1}^{N}
        w(t_i)
        \left(\frac{\partial\hat{\vx}^{i}}{\partial\theta}\right)^{\!\top}
        \Delta s_i
        \right].
    \label{eq:df_dmd_gradient}
\end{equation}
The teacher thus enters \cref{eq:df_dmd_gradient} only through
$[s_{\mathrm{real}}]_i$, a score conditioned on the context of block~$i$.
Sampling and distillation differ precisely in this context.
In sampling, the context must be clean ($\vt^{<i}=\vzero$): the teacher treats
its own imperfect frames as ground truth, so errors compound block by block.
In distillation, the context is the student's rollout, and its noise level is
a design choice rather than a requirement.
Consequently, the propagation of the teacher's sampling errors to the student
can be substantially mitigated if distillation avoids scoring under this clean
yet drifted context.
Whether another context noise level is available depends on how the teacher
is trained, and this is precisely where TF and DF teachers differ.

\paragraph{Teacher-Forcing teachers.}
A TF teacher is trained solely on clean context ($\vt^{<i}=\vzero$) and thus
defines a single conditional score
$[s_{\mathrm{real}}]_i=s_{\mathrm{real}}(\hat{\vx}_{t_i}^i,t_i\mid\hat{\vx}^{<i},\vzero,\vc^{\leq i})$.
Distillation therefore has no alternative to the clean-context score used in
sampling: the context now comes from the student's rollout rather than the
teacher's own, but is equally imperfect.
We call this scheme \emph{TF-style DMD}: as in teacher-forcing training, each
noisy target block is scored against a separate context sequence, which is
clean here but can in general be held at any fixed noise level.
As expected, the student closely tracks its teacher (\wbTFclean{} vs.\
\wbTFteacher{} in \cref{fig:sec3_wbench}); the small gain likely stems from
the mode-seeking behavior of the reverse KL objective.

\paragraph{Diffusion-Forcing teachers.}
A DF teacher is instead trained under random context noise
(\cref{eq:diffusion_forcing}) and thus learns a family of conditional scores
$[s_{\mathrm{real}}]_i=s_{\mathrm{real}}(\hat{\vx}_{t_i}^{i},t_i\mid\hat{\vx}_{\vt}^{<i},\vt^{<i},\vc^{\leq i})$
indexed by the context noise levels $\vt^{<i}$.
Its sampling still conditions on clean context and accumulates errors
(\cref{fig:teacher_student_qualitative}, top), but distillation can now query
any member of this family, leading to two distinct regimes:
\begin{itemize}[leftmargin=*, nosep, topsep=3pt, itemsep=3pt]
    \item \textbf{Clean-context scoring:}
          Applying TF-style DMD with clean context to a DF teacher queries the
          same fragile score used in sampling.
          The student again tracks its teacher and even falls slightly below
          it (\wbDFclean{} vs.\ \wbDFteacher), as high-noise-shifted DF
          pretraining leaves the clean-context score relatively under-trained.
    \item \textbf{High-noise context scoring:}
          Diffusing the context with high noise largely masks rollout errors and
          queries the teacher where it is well trained, thereby
          \emph{decoupling distillation from sampling}.
          The student should then no longer be bound by its teacher's sampling
          quality.
\end{itemize}

Prior Causal DMD, represented by CMD~\cite{bandyopadhyay2026cmd}, stays close
to the first regime.
CMD argues that each block should be scored under the context in which the
student generated it, and thus adopts scoring against a low-noise copy of the
rollout (\emph{Full CMD}) over a single noisy sequence (\emph{Base CMD}) as
its main method (details in Appendix~\ref{app:tf_style_dmd}).
Our findings point the opposite way: scoring under high-noise context, as
realized by \method in \cref{sec:method}, lifts the student of the \emph{same} DF teacher to
\wbDFours{}, outperforming its teacher and distillation with either
bidirectional or causal teachers, while avoiding the teacher's rollout drift
(\cref{fig:teacher_student_qualitative}, bottom).

\section{Causal Self Forcing}
\label{sec:method}

Motivated by \cref{sec:rethink}, we propose \emph{Causal Self Forcing (\method)},
which pairs Self Forcing's on-policy rollout~\cite{huang2025selfforcing} with \emph{causal} Diffusion-Forcing score
distillation. As shown in \cref{fig:df_dmd_comparison}, each generator update
consists of two steps:

\begin{figure*}[t]
    \centering
    \includegraphics[width=0.96\textwidth]{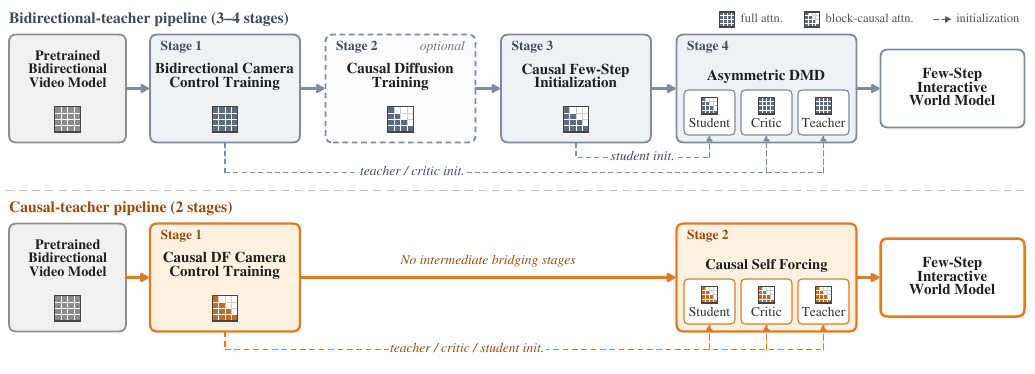}
    \caption{\textbf{Comparison of interactive world model training pipelines.}
        \textbf{Bidirectional-teacher pipeline (3--4 stages, top):} bidirectional
        camera control training (Stage~1) is followed by optional causal diffusion
        training (Stage~2), causal few-step initialization (Stage~3), and
        asymmetric DMD with a bidirectional teacher and critic (Stage~4).
        \textbf{Causal-teacher pipeline (2 stages, bottom):} causal Diffusion
        Forcing camera control training (Stage~1) produces a single checkpoint that
        directly initializes the causal student, critic, and frozen teacher for
        Causal Self Forcing (Stage~2).}
    \label{fig:two_stage_pipeline}
\end{figure*}

\begin{enumerate}[leftmargin=*, nosep, topsep=3pt, itemsep=3pt]
    \item \textbf{Self-forcing student rollout} (\cref{fig:df_dmd_comparison},
          panel~1)\textbf{:}
          Given a clean initial frame and blockwise controls $\vc^{1:N}$, the few-step
          causal student $G_\theta$ autoregressively generates
          $\hat{\vx}^{1:N}=(\hat{\vx}^1,\ldots,\hat{\vx}^N)$ on its own history,
          conditioning each block $\hat{\vx}^i$ on clean KV caches to avoid exposure bias.
    \item \textbf{Causal DF-style score distillation}
          (\cref{fig:df_dmd_comparison}, panels~2--4)\textbf{:}
          Rather than conditioning the teacher on clean or fixed-low-noise context,
          we sample a timestep $t$ from the same high-noise-shifted
          distribution as DF pretraining (\cref{sec:background_ar}), share it
          across blocks as in Self Forcing, $\vt=(t,\ldots,t)$, and draw
          Gaussian noises
          $\vepsilon^{1:N}\sim\mathcal{N}(\vzero,\mI)$ to perturb the rollout into a
          single noisy sequence $\hat{\vx}_{\vt}^{1:N}$.
          Just as in DF pretraining, each block
          $\hat{\vx}_{t_i}^i$ serves both as the scoring target for block $i$ and as
          part of the noisy causal context $\hat{\vx}_{\vt}^{<j}$ for subsequent blocks
          $j>i$ (dark and light cells of the attention masks in
          \cref{fig:df_dmd_comparison}), keeping the sequence length at $1\times$ tokens rather than $2\times$
          as in TF-style DMD~\cite{bandyopadhyay2026cmd}.
          A single forward pass of the frozen DF teacher $s_{\mathrm{real}}$ and
          the DF critic $s_{\mathrm{fake}}$ (held frozen in this update) under a
          block-causal mask instantiates
          the per-block score difference $\Delta s_i$ as
          \begin{equation}
              \begin{aligned}
                  \Delta s_i
                   & = s_{\mathrm{fake}}(\hat{\vx}_{t_i}^{i},t_i\mid\hat{\vx}_{\vt}^{<i},\vt^{<i},\vc^{\leq i})        \\
                   & \quad - s_{\mathrm{real}}(\hat{\vx}_{t_i}^{i},t_i\mid\hat{\vx}_{\vt}^{<i},\vt^{<i},\vc^{\leq i}),
              \end{aligned}
              \label{eq:df_dmd_score}
          \end{equation}
          which updates $G_\theta$ via \cref{eq:df_dmd_gradient}.
\end{enumerate}
Generator updates alternate with critic updates, in which $G_\theta$ is held
fixed and the DF critic $s_{\mathrm{fake}}$ is trained on detached rollouts
$\hat{\vx}^{1:N}$ with the DF loss in \cref{eq:diffusion_forcing}, so that it
tracks the score of the current student distribution.
The full training procedure is summarized in Appendix~\ref{app:algorithm}.

\paragraph{Reliable teacher scoring.}
\method queries the DF teacher where its score is most reliable.
As \cref{sec:rethink} shows, this requires high context noise, which \method
obtains by simply reusing the high-noise-shifted timestep distribution of DF
pretraining.
The teacher is thus queried in the regime where it was
most heavily supervised.

\paragraph{Causal supervision.}
Once high-noise scoring makes the causal teacher reliable, \method also
eliminates the conditioning mismatch of asymmetric
DMD~\cite{huang2025selfforcing}, where the block-$i$ score conditions on
future frames $\hat{\vx}_t^{>i}$ and future controls $\vc^{>i}$ unavailable
to a causal student (Appendix~\ref{app:appendix_bi_dmd}).
\method instead supervises each block solely with causal context and current
controls $(\hat{\vx}_{\vt}^{<i},\vc^{\leq i})$.

\paragraph{Direct two-stage training pipeline.}
Sharing the block-causal architecture across teacher, critic, and student
further simplifies world model training into two
stages~\cite{bandyopadhyay2026cmd}.
Because a causal student cannot be initialized from a bidirectional teacher,
existing pipelines~\cite{huang2026solarwm,zhao2026minwm} require three to four
stages with intermediate causal diffusion and few-step
initialization~\cite{zhu2026causalforcing,zhao2026causalforcingpp,zheng2026causalrcm}
(\cref{fig:two_stage_pipeline}, top), increasing training cost and progressively
degrading pretrained visual priors.
By contrast (\cref{fig:two_stage_pipeline}, bottom), \emph{Stage~1} jointly
trains camera control and causal DF on the pretrained video backbone, and
\emph{Stage~2} directly initializes $s_{\mathrm{real}}$, $s_{\mathrm{fake}}$,
and $G_\theta$ from the same Stage~1 checkpoint for \method distillation.

\section{Experiments}
\label{sec:experiments}

\begin{table*}[!tp]
    \centering
    \small
    \setlength{\tabcolsep}{3pt}
    \caption{\textbf{Quantitative comparisons on the WBench Navigation split~\cite{ying2026wbench}.} We report the overall score and the five WBench
        dimensions. Full CMD denotes the best-performing CMD configuration
        reported in~\cite{bandyopadhyay2026cmd}. SF denotes the released
        SolarWM-5B checkpoint~\cite{huang2026solarwm}, distilled with Self
        Forcing under a bidirectional teacher. User Rating is the average rank
        in our user study.}
    \label{tab:main_world_models}
    \begin{tabular*}{\textwidth}{@{\extracolsep{\fill}}lccccccc@{}}
        \toprule
        Model & Overall$\uparrow$ & Quality$\uparrow$ & Consistency$\uparrow$ & Interaction$\uparrow$ & Setting$\uparrow$ & Physical$\uparrow$ & User Rating$\downarrow$ \\
        \midrule
        \multicolumn{8}{@{}l}{Base Model: Wan2.2-5B} \\
        \quad+~CausVid~\cite{yin2025causvid} & 70.9 & 75.0 & 76.6 & 78.2 & 76.9 & 48.0 & 3.06 \\
        \quad+~Full CMD~\cite{bandyopadhyay2026cmd} & 72.3 & 76.6 & 78.1 & 80.0 & 76.6 & 50.1 & 2.94 \\
        \quad+~SF (SolarWM-5B~\cite{huang2026solarwm}) & 72.8 & 76.2 & 76.6 & 81.0 & 75.5 & 54.8 & 2.33 \\
        \quad+~\method (ours) & \textbf{75.7} & \textbf{78.5} & \textbf{81.7} & \textbf{82.8} & \textbf{77.1} & \textbf{58.6} & \textbf{1.67} \\
        \bottomrule
    \end{tabular*}
\end{table*}

\begin{table*}[!tp]
    \centering
    \small
    \setlength{\tabcolsep}{3pt}
    \caption{\textbf{Quantitative comparisons on the SANA-WM benchmark~\cite{zhu2026sanawm}.}
        All methods generate 60\,s videos at 480p. VBench is the aggregated
        VBench~\cite{huang2024vbench} Overall score; RotErr, TransErr, and CamMC
        measure camera-control errors against the ground-truth trajectory;
        PSNR, SSIM, and LPIPS measure revisit consistency between
        generated frames at nearly identical camera viewpoints; $\Delta$IQ
        measures temporal quality degradation as the imaging-quality drop from
        the first to the last 10\,s window.
        \textbf{Bold}: best; \underline{underline}: second best.}
    \label{tab:sana_wm}
    \begin{tabular*}{\textwidth}{@{\extracolsep{\fill}}lccccccccc@{}}
        \toprule
        & & \multicolumn{4}{c}{Trajectory and Visual Quality} & \multicolumn{3}{c}{Revisit Consistency} & \multicolumn{1}{c}{Temporal Quality} \\
        \cmidrule(lr){3-6} \cmidrule(lr){7-9} \cmidrule(l){10-10}
        Model & Params & VBench$\uparrow$ & RotErr$\downarrow$ & TransErr$\downarrow$ & CamMC$\downarrow$
        & PSNR$\uparrow$ & SSIM$\uparrow$ & LPIPS$\downarrow$ & $\Delta$IQ$\downarrow$ \\
        \midrule
        \multicolumn{10}{@{}l}{\emph{Simple-trajectory split}} \\
        HY-WorldPlay~\cite{sun2025worldplay} & 8B & 77.46 & \textbf{9.27} & \textbf{1.79} & \textbf{1.82} & \underline{14.31} & \underline{0.3388} & \textbf{0.4490} & 6.48 \\
        LingBot-World-Fast~\cite{gao2026lingbot} & 14B & \underline{78.34} & 27.49 & 2.50 & 2.70 & 12.59 & 0.3156 & 0.5853 & \textbf{4.43} \\
        \multicolumn{10}{@{}l}{Base Model: Wan2.2-5B} \\
        \quad+~CausVid~\cite{yin2025causvid} & 5B & 68.91 & 26.75 & 2.61 & 2.79 & 11.94 & 0.1738 & 0.5992 & 15.19 \\
        \quad+~SF (SolarWM-5B~\cite{huang2026solarwm}) & 5B & 73.99 & 20.64 & 2.38 & 2.50 & \textbf{14.90} & \textbf{0.4100} & \underline{0.5334} & 24.31 \\
        \quad+~Full CMD~\cite{bandyopadhyay2026cmd} & 5B & 76.93 & 25.21 & 2.42 & 2.59 & 12.49 & 0.2475 & 0.5867 & 6.34 \\
        \quad+~\method (ours) & 5B & \textbf{78.40} & \underline{9.88} & \underline{1.81} & \underline{1.85} & 13.21 & 0.2656 & 0.5371 & \underline{5.19} \\
        \midrule
        \multicolumn{10}{@{}l}{\emph{Hard-trajectory split}} \\
        HY-WorldPlay~\cite{sun2025worldplay} & 8B & \underline{78.67} & \underline{23.28} & \underline{1.90} & \underline{2.09} & \underline{13.29} & 0.2734 & \textbf{0.4784} & 4.21 \\
        LingBot-World-Fast~\cite{gao2026lingbot} & 14B & 78.39 & 42.89 & 2.55 & 2.89 & 12.61 & \underline{0.3035} & 0.5774 & \textbf{2.60} \\
        \multicolumn{10}{@{}l}{Base Model: Wan2.2-5B} \\
        \quad+~CausVid~\cite{yin2025causvid} & 5B & 70.18 & 29.86 & 2.18 & 2.42 & 11.96 & 0.2186 & 0.5383 & 9.67 \\
        \quad+~SF (SolarWM-5B~\cite{huang2026solarwm}) & 5B & 75.07 & 28.61 & 2.11 & 2.33 & \textbf{15.24} & \textbf{0.3901} & \underline{0.5243} & 19.67 \\
        \quad+~Full CMD~\cite{bandyopadhyay2026cmd} & 5B & 77.26 & 29.05 & 2.08 & 2.32 & 12.69 & 0.2591 & 0.5855 & 6.11 \\
        \quad+~\method (ours) & 5B & \textbf{79.40} & \textbf{19.86} & \textbf{1.84} & \textbf{1.99} & 13.11 & 0.2712 & 0.5339 & \underline{3.42} \\
        \bottomrule
    \end{tabular*}
\end{table*}

\subsection{Setup}
\label{sec:exp_setup}
\paragraph{Model.}
We build on Wan2.2-TI2V-5B~\cite{wan2025} and, following
SolarWM-5B~\cite{huang2026solarwm}, adopt fused-PRoPE camera conditioning.
The teacher, critic, and student share the same block-causal architecture,
with three latent frames per autoregressive block.
The few-step student generates each block in 4 denoising steps, followed by
one extra forward pass to update its clean KV cache.

\paragraph{Training.}
We train on the dataset of SolarWM-5B using 81-frame clips (about 5\,s) at
$864\times480$ resolution and 16 fps.
Stage~1 trains the causal DF teacher for 40k steps, a budget comparable to
the 30k-step bidirectional-teacher training plus causal initialization of
SolarWM-5B.
Stage~2 initializes the teacher, critic, and student from this checkpoint and
converges within 1k distillation steps.
For the baselines, SF is the released SolarWM-5B checkpoint, while CausVid
and Full CMD are trained by us under the same setup.
SF, Full CMD, and \method use Self Gradient Forcing
(SGF)~\cite{zhuang2026self}; CausVid does not roll out the student during
training and thus does not use SGF.

\paragraph{Evaluation.}
We adopt WBench~\cite{ying2026wbench} as our primary evaluation benchmark and
use its navigation split, which assesses short-horizon generation
(8--24\,s).
To assess minute-scale generation, we further adopt the SANA-WM
benchmark~\cite{zhu2026sanawm}, which pairs 80 initial scenes with simple and
hard camera trajectories and generates a 60\,s video for each.
It reports camera-control accuracy, VBench~\cite{huang2024vbench} quality,
revisit consistency, and imaging-quality degradation over time; for a fair
comparison, we only include models generating at 480p.
Additionally, we conduct a user study with 10 participants on 20 groups of
videos, where participants rank the overall quality of four
Wan2.2-5B-based methods.

More experimental details are provided in
Appendix~\ref{app:appendix_experiments}.

\subsection{Quantitative Results}
\label{sec:exp_quantitative}

\paragraph{Performance comparison with existing models.}
\label{sec:exp_user}
We analyze the results from the short-horizon and the long-horizon
perspectives.

\emph{Short-horizon generation.}
\cref{tab:main_world_models} reports the WBench results, where all methods
share the same Wan2.2-5B base model and differ only in how the few-step
autoregressive student is obtained. \method achieves the best overall score
and ranks first on all five dimensions, improving the overall score over the
standard practice SF (SolarWM-5B) from 72.8 to 75.7. The largest gains are on
consistency (81.7 vs.~78.1 for the second best) and physics compliance
(58.6 vs.~54.8). Human preference is consistent with these metrics: in the user study, \method
is ranked best on average (1.67), ahead of SF (2.33).

\begin{figure*}[!t]
    \centering
    \includegraphics[width=\textwidth]{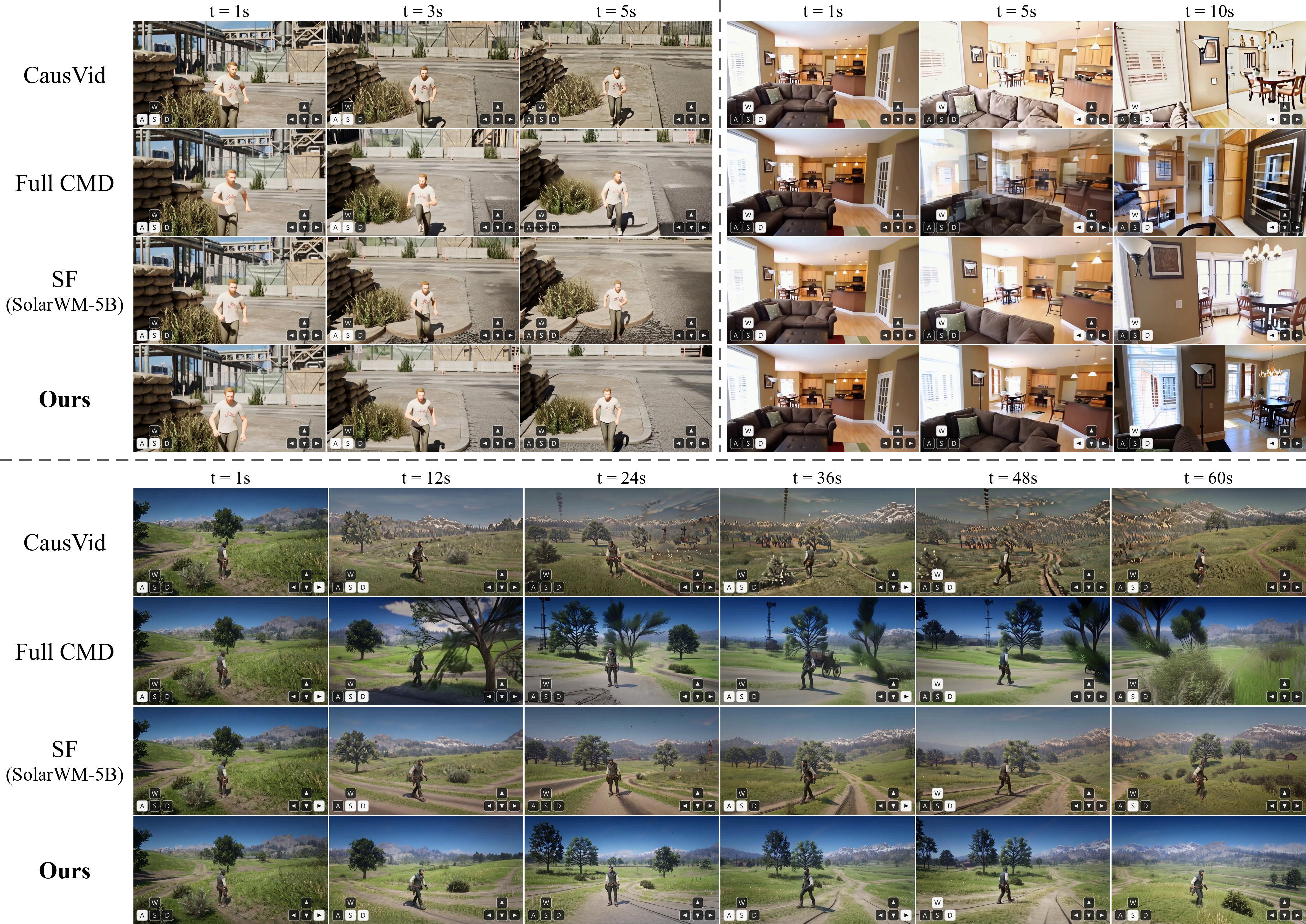}
    \caption{\textbf{Qualitative comparisons with existing methods.} All methods are
        built on Wan2.2-5B and conditioned on the same input image and camera
        trajectory; columns show frames at the indicated timestamps of
        5\,s and 10\,s rollouts (top) and a 60\,s rollout (bottom).
    }
    \label{fig:qualitative}
\end{figure*}

\emph{Long-horizon generation.}
\cref{tab:sana_wm} reports the 60\,s results on the SANA-WM benchmark, where
videos are over ten times longer than the training clips. At this horizon,
the advantage of \method over the other Wan2.2-5B-based methods becomes more
pronounced. It shows the least temporal quality degradation among them, with a
$\Delta$IQ of 5.19 and 3.42 on the two splits against 24.31 and 19.67 for SF,
and follows the camera trajectory far more accurately, more than halving the
rotation error on the simple split (9.88 vs.~20.64--26.75).

\method also compares favorably with HY-WorldPlay (8B) and LingBot-World-Fast
(14B), which are larger in both parameters and training scale. It attains the
highest VBench score among all methods on both splits. In camera control, it
is on par with HY-WorldPlay on the simple split, surpasses it on the hard
split, and is far more accurate than LingBot-World-Fast on both (rotation
error of 9.88 vs.~27.49 and 19.86 vs.~42.89). In temporal quality, its
$\Delta$IQ is lower than that of HY-WorldPlay and slightly higher than that of
LingBot-World-Fast (4.43 and 2.60).

The remaining gap lies in revisit consistency. HY-WorldPlay benefits from an
explicit memory mechanism that \method lacks, which is orthogonal to our
contribution. The high PSNR and SSIM of SF are largely a byproduct of its
degradation: its late frames become blurry, which inflates pixel-level
similarity, whereas its advantage on the perceptual LPIPS is marginal.

\paragraph{Ablation studies.}
\cref{sec:rethink} has analyzed the choice of the causal teacher and how it
scores the student.
Here we ablate the other core components of \method on WBench; further
analysis of causal teachers is provided in Appendix~\ref{app:teacher_analysis}.

\emph{On-policy rollout vs.~noised ground truth.}
In \cref{tab:ablation_rollout}, we keep the DF teacher, the causal critic, the
initialization, and the high-noise scoring of \method fixed, and only replace
the student's self-generated rollouts with noised ground truth during
distillation, following the causal-teacher setting of
CausVid~\cite{yin2025causvid}.
This variant reaches only 67.8, below even CausVid with a bidirectional
teacher (70.9 in \cref{tab:main_world_models}), consistent with the weak
causal-teacher results reported by CausVid.
Switching to on-policy rollouts raises the score to 75.7.
Together with \cref{sec:rethink}, where on-policy rollouts with clean-context
scoring reach only \wbDFclean{}, this shows that the two ingredients are
complementary: on-policy rollouts avoid exposure bias, whereas high-noise
scoring keeps the teacher's sampling errors from reaching the student, and
neither alone suffices.

\begin{table}[!t]
    \centering
    \footnotesize
    \setlength{\tabcolsep}{1.5pt}
    \caption{\textbf{Ablation on the student input during distillation (WBench
            navigation).} Both variants use the same DF teacher, critic, and
        initialization. Noised GT follows the causal-teacher setting of
        CausVid~\cite{yin2025causvid}. Qua.: quality, Con.: consistency, Int.: interaction,
        Set.: setting, Phy.: physical.}
    \label{tab:ablation_rollout}
    \begin{tabular*}{\columnwidth}{@{\extracolsep{\fill}}lcccccc@{}}
        \toprule
        Student input & Overall$\uparrow$ & Qua.$\uparrow$ & Con.$\uparrow$ & Int.$\uparrow$ & Set.$\uparrow$ & Phy.$\uparrow$ \\
        \midrule
        Noised GT (CausVid-style) & 67.8 & 73.1 & 74.3 & 72.8 & 74.8 & 43.8 \\
        Self-forcing rollout (\method) & \textbf{75.7} & \textbf{78.5} & \textbf{81.7} & \textbf{82.8} & \textbf{77.1} & \textbf{58.6} \\
        \bottomrule
    \end{tabular*}
\end{table}

\emph{Teacher training steps.}
\cref{tab:ablation_teacher_steps} distills students from DF teacher
checkpoints at different pretraining steps. The student improves steadily
with the teacher, from 59.3 at 10k steps to a peak of 75.7 at 40k steps, and
declines slightly afterwards. We attribute
this decline to the limited quality of the open-source training
data~\cite{huang2026solarwm}: once the camera and causal adaptation has
converged, further training mainly pulls the teacher away from the
pretrained prior toward the data, as reflected by the drop in the quality
score. 
Notably, the student of the 30k-step teacher already reaches 74.2 and
outperforms SF (72.8), whose bidirectional teacher is also trained for 30k
steps, so the gain of \method does not come from longer teacher training.

\begin{table}[!t]
    \centering
    \footnotesize
    \setlength{\tabcolsep}{1.5pt}
    \caption{\textbf{WBench navigation scores of students distilled with \method from DF teacher checkpoints at different pretraining steps.}
        Steps refer to teacher pretraining, not distillation.
        Metric abbreviations follow ~\cref{tab:ablation_rollout}.}
    \label{tab:ablation_teacher_steps}
    \begin{tabular*}{\columnwidth}{@{\extracolsep{\fill}}lcccccc@{}}
        \toprule
        Teacher steps & Overall$\uparrow$ & Qua.$\uparrow$ & Con.$\uparrow$ & Int.$\uparrow$ & Set.$\uparrow$ & Phy.$\uparrow$ \\
        \midrule
        10k & 59.3 & 61.9 & 70.8 & 70.5 & 59.7 & 33.5 \\
        20k & 72.7 & 76.7 & 79.7 & \textbf{83.5} & \textbf{77.8} & 45.7 \\
        30k & 74.2 & 78.2 & 80.3 & 81.6 & 77.1 & 53.6 \\
        40k & \textbf{75.7} & \textbf{78.5} & \textbf{81.7} & 82.8 & 77.1 & 58.6 \\
        50k & 75.0 & 77.7 & 80.6 & 81.9 & 76.3 & \textbf{58.9} \\
        60k & 74.5 & 77.6 & 80.1 & 80.4 & 76.9 & 57.6 \\
        \bottomrule
    \end{tabular*}
\end{table}

\subsection{Qualitative Results}
\label{sec:exp_qualitative}

\Cref{fig:qualitative} compares \method with CausVid, Full CMD, and SF
(SolarWM-5B) under the same input image and camera trajectory, and the
observations mirror the quantitative results.
At short horizons (top), all methods remain plausible within the 5\,s training
length, but differences already emerge at 10\,s: CausVid becomes overexposed
and loses fine details, and Full CMD exhibits ghosting at 5\,s and drifts to
an unrelated scene by 10\,s, whereas SF and \method preserve the room layout,
with \method retaining sharper details.
The gap widens in the 60\,s rollout (bottom). CausVid accumulates structural
artifacts from around 24\,s until the scene collapses, and Full CMD suffers
from severe motion blur and scene drift. SF keeps the scene layout but
gradually loses contrast and sharpness, ending with hazy and blurry frames,
which agrees with its large $\Delta$IQ in \cref{tab:sana_wm}. In contrast,
\method maintains a stable scene layout and consistent visual fidelity
throughout the full minute.

More qualitative results are provided in Appendix~\ref{app:qualitative}.

\section{Discussion}
\label{sec:discussion}
In this part, we discuss the broader implications of our results and possible directions for future research.
\paragraph{Importance of the causal teacher.}
Although a \method student can surpass its DF teacher (\cref{sec:rethink}),
the teacher remains the sole source of supervision: the student is trained to
match the distribution defined by the teacher's scores, so their quality
bounds what the student can reach.
\cref{tab:ablation_teacher_steps} confirms this dependence, as student
performance varies substantially with the teacher's training steps. We further observe
that a DF teacher trained on lower-quality data, \eg, model-generated videos,
yields a clearly worse student. A well-trained causal teacher and high-quality
training data therefore remain essential to \method.

\paragraph{Long video generation.}
Although \method is trained only on 5\,s clips, it generates stable 60\,s
videos (\cref{tab:sana_wm}). This ability builds on SGF~\cite{zhuang2026self},
which enables native extrapolation beyond the training length. The gains of \method on top of
it indicate that a better distillation objective and long-video techniques are
complementary. Long-horizon generation can be further improved by training on
rollouts longer than the training
clips~\cite{yang2025longlive,liu2025rolling,cui2025self,chen2026contextforcing}
or by training-free inference
strategies~\cite{yesiltepe2026infinity,yi2025deep}. These directions are
orthogonal to \method, and we leave their combination to future work.

\section{Conclusion}
\label{sec:conclusion}
We revisit the prevailing belief that causal teachers inherently yield weaker
students in DMD.
Since the teacher only scores student rollouts, the propagation of its
sampling errors can be substantially mitigated if it avoids scoring under a
clean yet drifted context.
Motivated by this insight, we propose Causal Self Forcing, which
lets a DF teacher score on-policy student rollouts under heavily noised
context.
Under strictly matched settings on Wan2.2-5B, the resulting few-step student
outperforms distillation with either bidirectional or causal teachers,
surpasses its own multi-step teacher, and in minute-long generation even
matches or exceeds substantially larger world models.
We hope these findings encourage the community to reconsider causal teachers
for autoregressive video distillation.

\section*{LLM Usage Statement}
We used Claude and Gemini for language editing and grammar
polishing throughout the manuscript. We used GPT as
coding assistants for implementing plotting/visualization scripts and
experiment utilities; all such code was reviewed, tested, and verified by the
authors. All technical content, ideas, experiments, and results were
produced and verified by the authors, who take full responsibility for the paper.

{
  \small
  \bibliographystyle{ieeenat_fullname}
  \bibliography{main}

\begin{thebibliography}{94}
\providecommand{\natexlab}[1]{#1}
\providecommand{\url}[1]{\texttt{#1}}
\expandafter\ifx\csname urlstyle\endcsname\relax
  \providecommand{\doi}[1]{doi: #1}\else
  \providecommand{\doi}{doi: \begingroup \urlstyle{rm}\Url}\fi

\bibitem[Alonso et~al.(2024)Alonso, Jelley, Micheli, Kanervisto, Storkey, Pearce, and Fleuret]{alonso2024diamond}
Eloi Alonso, Adam Jelley, Vincent Micheli, Anssi Kanervisto, Amos Storkey, Tim Pearce, and Fran{\c{c}}ois Fleuret.
\newblock Diffusion for world modeling: Visual details matter in {Atari}.
\newblock In \emph{Advances in Neural Information Processing Systems}, 2024.

\bibitem[Ball et~al.(2025)Ball, Bauer, Belletti, Brownfield, Ephrat, Fruchter, Gupta, Holsheimer, Holynski, Hron, Kaplanis, Limont, McGill, Oliveira, Parker-Holder, Perbet, Scully, Shar, Spencer, Tov, Villegas, Wang, Yung, Baetu, Berbel, Bridson, Bruce, Buttimore, Chakera, Chandra, Collins, Cullum, Damoc, Dasagi, Gazeau, Gbadamosi, Han, Hirst, Kachra, Kerley, Kjems, Knoepfel, Koriakin, Lo, Lu, Mehring, Moufarek, Nandwani, Oliveira, Pardo, Park, Pierson, Poole, Ran, Salimans, Sanchez, Saprykin, Shen, Sidhwani, Smith, Stanton, Tomlinson, Vijaykumar, Wang, Wingfield, Wong, Xu, Yew, Young, Zubov, Eck, Erhan, Kavukcuoglu, Hassabis, Gharamani, Hadsell, van~den Oord, Mosseri, Bolton, Singh, and Rockt{\"a}schel]{genie3}
Philip~J. Ball, Jakob Bauer, Frank Belletti, Bethanie Brownfield, Ariel Ephrat, Shlomi Fruchter, Agrim Gupta, Kristian Holsheimer, Aleksander Holynski, Jiri Hron, Christos Kaplanis, Marjorie Limont, Matt McGill, Yanko Oliveira, Jack Parker-Holder, Frank Perbet, Guy Scully, Jeremy Shar, Stephen Spencer, Omer Tov, Ruben Villegas, Emma Wang, Jessica Yung, Cip Baetu, Jordi Berbel, David Bridson, Jake Bruce, Gavin Buttimore, Sarah Chakera, Bilva Chandra, Paul Collins, Alex Cullum, Bogdan Damoc, Vibha Dasagi, Maxime Gazeau, Charles Gbadamosi, Woohyun Han, Ed Hirst, Ashyana Kachra, Lucie Kerley, Kristian Kjems, Eva Knoepfel, Vika Koriakin, Jessica Lo, Cong Lu, Zeb Mehring, Alex Moufarek, Henna Nandwani, Valeria Oliveira, Fabio Pardo, Jane Park, Andrew Pierson, Ben Poole, Helen Ran, Tim Salimans, Manuel Sanchez, Igor Saprykin, Amy Shen, Sailesh Sidhwani, Duncan Smith, Joe Stanton, Hamish Tomlinson, Dimple Vijaykumar, Luyu Wang, Piers Wingfield, Nat Wong, Keyang Xu, Christopher Yew, Nick Young, Vadim Zubov, Douglas Eck, Dumitru Erhan, Koray Kavukcuoglu, Demis Hassabis, Zoubin Gharamani, Raia Hadsell, A{\"a}ron van~den Oord, Inbar Mosseri, Adrian Bolton, Satinder Singh, and Tim Rockt{\"a}schel.
\newblock Genie 3: A new frontier for world models.
\newblock 2025.

\bibitem[Bandyopadhyay et~al.(2026)Bandyopadhyay, Ren, Huang, Wu, Cao, Li, Chu, Fidler, Song, and Wang]{bandyopadhyay2026cmd}
Hmrishav Bandyopadhyay, Xuanchi Ren, Zijian Huang, Jay~Zhangjie Wu, Tianshi Cao, Ruilong Li, Bryan Chu, Sanja Fidler, Yi-Zhe Song, and Zian Wang.
\newblock Context-matched distillation: Teacher causality for autoregressive video distillation.
\newblock \emph{arXiv preprint arXiv:2608.13391}, 2026.

\bibitem[Bao et~al.(2024)Bao, Xiang, Yue, He, Zhu, Zheng, Zhao, Liu, Wang, and Zhu]{bao2024vidu}
Fan Bao, Chendong Xiang, Gang Yue, Guande He, Hongzhou Zhu, Kaiwen Zheng, Min Zhao, Shilong Liu, Yaole Wang, and Jun Zhu.
\newblock Vidu: A highly consistent, dynamic and skilled text-to-video generator with diffusion models.
\newblock \emph{arXiv preprint arXiv:2405.04233}, 2024.

\bibitem[Blattmann et~al.(2023)Blattmann, Rombach, Ling, Dockhorn, Kim, Fidler, and Kreis]{blattmann2023align}
Andreas Blattmann, Robin Rombach, Huan Ling, Tim Dockhorn, Seung~Wook Kim, Sanja Fidler, and Karsten Kreis.
\newblock Align your latents: High-resolution video synthesis with latent diffusion models.
\newblock In \emph{Proceedings of the IEEE/CVF conference on computer vision and pattern recognition}, pages 22563--22575, 2023.

\bibitem[Bruce et~al.(2024)Bruce, Dennis, Edwards, Parker-Holder, Shi, Hughes, Lai, Mavalankar, Steigerwald, Apps, et~al.]{bruce2024genie}
Jake Bruce, Michael~D. Dennis, Ashley Edwards, Jack Parker-Holder, Yuge Shi, Edward Hughes, Matthew Lai, Aditi Mavalankar, Richie Steigerwald, Chris Apps, et~al.
\newblock Genie: Generative interactive environments.
\newblock In \emph{International Conference on Machine Learning}, 2024.

\bibitem[Chen et~al.(2024)Chen, Mart{\'\i}~Mons{\'o}, Du, Simchowitz, Tedrake, and Sitzmann]{chen2024diffusion}
Boyuan Chen, Diego Mart{\'\i}~Mons{\'o}, Yilun Du, Max Simchowitz, Russ Tedrake, and Vincent Sitzmann.
\newblock Diffusion forcing: Next-token prediction meets full-sequence diffusion.
\newblock In \emph{Advances in Neural Information Processing Systems}, 2024.

\bibitem[Chen et~al.(2025)Chen, Lin, Yang, Lin, Zhu, Fan, Zhang, Chen, Chen, Ma, et~al.]{chen2025skyreels}
Guibin Chen, Dixuan Lin, Jiangping Yang, Chunze Lin, Junchen Zhu, Mingyuan Fan, Hao Zhang, Sheng Chen, Zheng Chen, Chengcheng Ma, et~al.
\newblock {SkyReels-V2}: Infinite-length film generative model.
\newblock \emph{arXiv preprint arXiv:2504.13074}, 2025.

\bibitem[Chen et~al.(2023)Chen, Xia, He, Zhang, Cun, Yang, Xing, Liu, Chen, Wang, et~al.]{chen2023videocrafter1}
Haoxin Chen, Menghan Xia, Yingqing He, Yong Zhang, Xiaodong Cun, Shaoshu Yang, Jinbo Xing, Yaofang Liu, Qifeng Chen, Xintao Wang, et~al.
\newblock Videocrafter1: Open diffusion models for high-quality video generation.
\newblock \emph{arXiv preprint arXiv:2310.19512}, 2023.

\bibitem[Chen et~al.(2026{\natexlab{a}})Chen, Chen, Fu, Gong, Guo, Li, Li, Li, Li, Liu, et~al.]{chen2026zing}
Mingyang Chen, Shengdong Chen, Xiaoxiao Fu, Bosheng Gong, Haoyuan Guo, Bowen Li, Jiawen Li, Kejun Li, Tianpeng Li, Yin Liu, et~al.
\newblock Zing-0.5: Toward playable worlds with real-time joint action and text control.
\newblock \emph{arXiv preprint arXiv:2609.17909}, 2026{\natexlab{a}}.

\bibitem[Chen et~al.(2026{\natexlab{b}})Chen, Wei, Sun, Shen, Nie, Zou, Zhang, Yang, and Chen]{chen2026contextforcing}
Shuo Chen, Cong Wei, Sun Sun, Tiancheng Shen, Ping Nie, Kai Zou, Ge Zhang, Ming-Hsuan Yang, and Wenhu Chen.
\newblock Context forcing: Consistent autoregressive video generation with long context.
\newblock \emph{arXiv preprint arXiv:2602.06028}, 2026{\natexlab{b}}.

\bibitem[Cui et~al.(2025)Cui, Wu, Li, Yang, Li, Wang, Bai, Ban, and Hsieh]{cui2025self}
Justin Cui, Jie Wu, Ming Li, Tao Yang, Xiaojie Li, Rui Wang, Andrew Bai, Yuanhao Ban, and Cho-Jui Hsieh.
\newblock {Self-Forcing++}: Towards minute-scale high-quality video generation.
\newblock \emph{arXiv preprint arXiv:2510.02283}, 2025.

\bibitem[Cui et~al.(2026)Cui, Wu, Li, Yang, Li, Wang, Bai, Ban, and Hsieh]{cui2026lol}
Justin Cui, Jie Wu, Ming Li, Tao Yang, Xiaojie Li, Rui Wang, Andrew Bai, Yuanhao Ban, and Cho-Jui Hsieh.
\newblock Lol: Longer than longer, scaling video generation to hour.
\newblock \emph{arXiv preprint arXiv:2601.16914}, 2026.

\bibitem[Deng et~al.(2024)Deng, Pan, Diao, Luo, Cui, Lu, Shan, Qi, and Wang]{deng2024autoregressive}
Haoge Deng, Ting Pan, Haiwen Diao, Zhengxiong Luo, Yufeng Cui, Huchuan Lu, Shiguang Shan, Yonggang Qi, and Xinlong Wang.
\newblock Autoregressive video generation without vector quantization.
\newblock \emph{arXiv preprint arXiv:2412.14169}, 2024.

\bibitem[{DreamX Team} et~al.(2026){DreamX Team}, Bai, Chen, Chu, Dang, Dou, Gao, Gu, Hong, Lei, Li, Li, Lin, Shi, Song, Sun, Tang, Tian, Wang, Wu, Zhang, Zhang, and Zhu]{team2026dreamx}
{DreamX Team}, Yancheng Bai, Rui Chen, Xiangxiang Chu, Rujing Dang, Hao Dou, Bingjie Gao, Qiwen Gu, Siyu Hong, Jiachen Lei, Geng Li, Jifan Li, Ruimin Lin, Qingfeng Shi, Bingze Song, Lei Sun, Jing Tang, Ruitian Tian, Jun Wang, Jiahong Wu, Pengfei Zhang, Shen Zhang, and Jiashu Zhu.
\newblock {DreamX-World 1.0}: A general-purpose interactive world model.
\newblock \emph{arXiv preprint arXiv:2606.16993}, 2026.

\bibitem[Feng et~al.(2026)Feng, Cui, Ban, and Hsieh]{feng2026one}
Jiaqi Feng, Justin Cui, Yuanhao Ban, and Cho-Jui Hsieh.
\newblock One-forcing: Towards stable one-step autoregressive video generation.
\newblock \emph{arXiv preprint arXiv:2605.23458}, 2026.

\bibitem[Gao et~al.(2024)Gao, Shi, Zhang, Wang, Xiao, and Chen]{gao2024ca2}
Kaifeng Gao, Jiaxin Shi, Hanwang Zhang, Chunping Wang, Jun Xiao, and Long Chen.
\newblock {CA2-VDM}: Efficient autoregressive video diffusion model with causal generation and cache sharing.
\newblock \emph{arXiv preprint arXiv:2411.16375}, 2024.

\bibitem[Gao et~al.(2026)Gao, Wang, Zhu, Chen, Liu, Bai, Wang, Yuan, Wang, Lu, et~al.]{gao2026lingbot2}
Zelin Gao, Qiuyu Wang, Jiapeng Zhu, Jingye Chen, Zichen Liu, Qingyan Bai, Jiahao Wang, Yufeng Yuan, Hanlin Wang, Yichong Lu, et~al.
\newblock Infinite worlds with versatile interactions.
\newblock \emph{arXiv preprint arXiv:2607.07534}, 2026.

\bibitem[Gu et~al.(2026)Gu, Fang, Jiang, Mao, Han, Cai, and Shou]{gu2026anyflow}
Yuchao Gu, Guian Fang, Yuxin Jiang, Weijia Mao, Song Han, Han Cai, and Mike~Zheng Shou.
\newblock {AnyFlow}: Any-step video diffusion model with on-policy flow map distillation.
\newblock \emph{arXiv preprint arXiv:2605.13724}, 2026.

\bibitem[Guo et~al.(2025{\natexlab{a}})Guo, Ye, He, Wu, Jiang, Pearce, and Bian]{guo2025mineworld}
Junliang Guo, Yang Ye, Tianyu He, Haoyu Wu, Yushu Jiang, Tim Pearce, and Jiang Bian.
\newblock {MineWorld}: A real-time and open-source interactive world model on {Minecraft}.
\newblock \emph{arXiv preprint arXiv:2504.08388}, 2025{\natexlab{a}}.

\bibitem[Guo et~al.(2025{\natexlab{b}})Guo, Yang, He, Zhao, Wei, Yang, Huang, and Lin]{guo2025end}
Yuwei Guo, Ceyuan Yang, Hao He, Yang Zhao, Meng Wei, Zhenheng Yang, Weilin Huang, and Dahua Lin.
\newblock End-to-end training for autoregressive video diffusion via self-resampling.
\newblock \emph{arXiv preprint arXiv:2512.15702}, 2025{\natexlab{b}}.

\bibitem[HaCohen et~al.(2024)HaCohen, Chiprut, Brazowski, Shalem, Moshe, Richardson, Levin, Shiran, Zabari, Gordon, et~al.]{hacohen2024ltx}
Yoav HaCohen, Nisan Chiprut, Benny Brazowski, Daniel Shalem, Dudu Moshe, Eitan Richardson, Eran Levin, Guy Shiran, Nir Zabari, Ori Gordon, et~al.
\newblock {LTX-Video}: Realtime video latent diffusion.
\newblock \emph{arXiv preprint arXiv:2501.00103}, 2024.

\bibitem[He et~al.(2024)He, Xu, Guo, Wetzstein, Dai, Li, and Yang]{he2024cameractrl}
Hao He, Yinghao Xu, Yuwei Guo, Gordon Wetzstein, Bo Dai, Hongsheng Li, and Ceyuan Yang.
\newblock {CameraCtrl}: Enabling camera control for text-to-video generation.
\newblock \emph{arXiv preprint arXiv:2404.02101}, 2024.

\bibitem[He et~al.(2025)He, Peng, Liu, Wang, Zhang, Cui, Kang, Jiang, An, Ren, et~al.]{he2025matrixgame2}
Xianglong He, Chunli Peng, Zexiang Liu, Boyang Wang, Yifan Zhang, Qi Cui, Fei Kang, Biao Jiang, Mengyin An, Yangyang Ren, et~al.
\newblock {Matrix-Game 2.0}: An open-source, real-time, and streaming interactive world model.
\newblock \emph{arXiv preprint arXiv:2508.13009}, 2025.

\bibitem[He et~al.(2022)He, Yang, Zhang, Shan, and Chen]{he2022latent}
Yingqing He, Tianyu Yang, Yong Zhang, Ying Shan, and Qifeng Chen.
\newblock Latent video diffusion models for high-fidelity long video generation.
\newblock \emph{arXiv preprint arXiv:2211.13221}, 2022.

\bibitem[Ho et~al.(2020)Ho, Jain, and Abbeel]{ho2020denoising}
Jonathan Ho, Ajay Jain, and Pieter Abbeel.
\newblock Denoising diffusion probabilistic models.
\newblock In \emph{Advances in Neural Information Processing Systems}, pages 6840--6851, 2020.

\bibitem[Ho et~al.(2022)Ho, Chan, Saharia, Whang, Gao, Gritsenko, Kingma, Poole, Norouzi, Fleet, et~al.]{ho2022imagen}
Jonathan Ho, William Chan, Chitwan Saharia, Jay Whang, Ruiqi Gao, Alexey Gritsenko, Diederik~P Kingma, Ben Poole, Mohammad Norouzi, David~J Fleet, et~al.
\newblock Imagen video: High definition video generation with diffusion models.
\newblock \emph{arXiv preprint arXiv:2210.02303}, 2022.

\bibitem[Hong et~al.(2022)Hong, Ding, Zheng, Liu, and Tang]{hong2022cogvideo}
Wenyi Hong, Ming Ding, Wendi Zheng, Xinghan Liu, and Jie Tang.
\newblock Cogvideo: Large-scale pretraining for text-to-video generation via transformers.
\newblock \emph{arXiv preprint arXiv:2205.15868}, 2022.

\bibitem[Hu et~al.(2026)Hu, Wei, Huang, Huang, Wang, Yang, Wang, Zhu, Yang, Liu, et~al.]{hu2026zyt}
Boni Hu, Xiong Wei, Haoming Huang, Yong Huang, Chenbo Wang, Yi Yang, Jiancheng Wang, Ruicheng Zhu, Zhimin Yang, Guanglai Liu, et~al.
\newblock Zyt-world: A real-time controllable world model for closed-loop autonomous-driving simulation.
\newblock \emph{arXiv preprint arXiv:2609.21712}, 2026.

\bibitem[Huang et~al.(2026{\natexlab{a}})Huang, Fang, Qian, Kong, Zhao, Huang, Du, Zhang, Cui, Gu, Chen, Hu, He, Shi, Tian, Wang, Shou, and Jiang]{huang2026solarwm}
Junchao Huang, Guian Fang, Shengju Qian, Xianghao Kong, Zhuoran Zhao, Wei Huang, Yihua Du, Zixin Zhang, Justin Cui, Yuchao Gu, Yukang Chen, Xinting Hu, Tianyu He, Shaoshuai Shi, Zhuotao Tian, Xin Wang, Mike~Zheng Shou, and Li Jiang.
\newblock Solarwm: Open data and scalable training for long-horizon video world models.
\newblock \emph{arXiv preprint arXiv:2609.02886}, 2026{\natexlab{a}}.

\bibitem[Huang et~al.(2026{\natexlab{b}})Huang, Ye, Hu, He, Zhang, Shi, Bian, and Jiang]{huang2026live}
Junchao Huang, Ziyang Ye, Xinting Hu, Tianyu He, Guiyu Zhang, Shaoshuai Shi, Jiang Bian, and Li Jiang.
\newblock Live: Long-horizon interactive video world modeling.
\newblock \emph{arXiv preprint arXiv:2602.03747}, 2026{\natexlab{b}}.

\bibitem[Huang et~al.(2025{\natexlab{a}})Huang, Li, He, Zhou, and Shechtman]{huang2025selfforcing}
Xun Huang, Zhengqi Li, Guande He, Mingyuan Zhou, and Eli Shechtman.
\newblock Self forcing: Bridging the train-test gap in autoregressive video diffusion.
\newblock \emph{arXiv preprint arXiv:2506.08009}, 2025{\natexlab{a}}.

\bibitem[Huang et~al.(2025{\natexlab{b}})Huang, Guo, Wu, Zhang, Huang, Gan, Liu, Zhao, Chen, Liu, et~al.]{huang2025live}
Yubo Huang, Hailong Guo, Fangtai Wu, Shifeng Zhang, Shijie Huang, Qijun Gan, Lin Liu, Sirui Zhao, Enhong Chen, Jiaming Liu, et~al.
\newblock Live avatar: Streaming real-time audio-driven avatar generation with infinite length.
\newblock \emph{arXiv preprint arXiv:2512.04677}, 2025{\natexlab{b}}.

\bibitem[Huang et~al.(2024)Huang, He, Yu, Zhang, Si, Jiang, Zhang, Wu, Jin, Chanpaisit, Wang, Chen, Wang, Lin, Qiao, and Liu]{huang2024vbench}
Ziqi Huang, Yinan He, Jiashuo Yu, Fan Zhang, Chenyang Si, Yuming Jiang, Yuanhan Zhang, Tianxing Wu, Qingyang Jin, Nattapol Chanpaisit, Yaohui Wang, Xinyuan Chen, Limin Wang, Dahua Lin, Yu Qiao, and Ziwei Liu.
\newblock {VBench}: Comprehensive benchmark suite for video generative models.
\newblock In \emph{CVPR}, 2024.

\bibitem[Jin et~al.(2024)Jin, Sun, Li, Xu, Jiang, Zhuang, Huang, Song, Mu, and Lin]{jin2024pyramidal}
Yang Jin, Zhicheng Sun, Ningyuan Li, Kun Xu, Hao Jiang, Nan Zhuang, Quzhe Huang, Yang Song, Yadong Mu, and Zhouchen Lin.
\newblock Pyramidal flow matching for efficient video generative modeling.
\newblock \emph{arXiv preprint arXiv:2410.05954}, 2024.

\bibitem[Junyi~Chen(2025)]{infinite-forcing}
Xianglong~He Junyi~Chen, Zhoujie~Fu.
\newblock Infinite-forcing: Towards infinite-long video generation, 2025.
\newblock https://github.com/SOTAMak1r/Infinite-Forcing.

\bibitem[Kondratyuk et~al.(2023)Kondratyuk, Yu, Gu, Lezama, Huang, Schindler, Hornung, Birodkar, Yan, Chiu, et~al.]{kondratyuk2023videopoet}
Dan Kondratyuk, Lijun Yu, Xiuye Gu, Jos{\'e} Lezama, Jonathan Huang, Grant Schindler, Rachel Hornung, Vighnesh Birodkar, Jimmy Yan, Ming-Chang Chiu, et~al.
\newblock Videopoet: A large language model for zero-shot video generation.
\newblock \emph{arXiv preprint arXiv:2312.14125}, 2023.

\bibitem[Kong et~al.(2024)Kong, Tian, Zhang, Min, Dai, Zhou, Xiong, Li, Wu, Zhang, et~al.]{kong2024hunyuanvideo}
Weijie Kong, Qi Tian, Zijian Zhang, Rox Min, Zuozhuo Dai, Jin Zhou, Jiangfeng Xiong, Xin Li, Bo Wu, Jianwei Zhang, et~al.
\newblock {HunyuanVideo}: A systematic framework for large video generative models.
\newblock \emph{arXiv preprint arXiv:2412.03603}, 2024.

\bibitem[Li et~al.(2025)Li, Yi, Liu, Gao, Ma, and Kanazawa]{li2026prope}
Ruilong Li, Brent Yi, Junchen Liu, Hang Gao, Yi Ma, and Angjoo Kanazawa.
\newblock Cameras as relative positional encoding.
\newblock In \emph{Advances in Neural Information Processing Systems}, 2025.

\bibitem[Lin et~al.(2024)Lin, Ge, Cheng, Li, Zhu, Wang, He, Ye, Yuan, Chen, et~al.]{lin2024open}
Bin Lin, Yunyang Ge, Xinhua Cheng, Zongjian Li, Bin Zhu, Shaodong Wang, Xianyi He, Yang Ye, Shenghai Yuan, Liuhan Chen, et~al.
\newblock Open-sora plan: Open-source large video generation model.
\newblock \emph{arXiv preprint arXiv:2412.00131}, 2024.

\bibitem[Lin et~al.(2025)Lin, Yang, He, Jiang, Ren, Xia, Zhao, Xiao, and Jiang]{lin2026autoregressive}
Shanchuan Lin, Ceyuan Yang, Hao He, Jianwen Jiang, Yuxi Ren, Xin Xia, Yang Zhao, Xuefeng Xiao, and Lu Jiang.
\newblock Autoregressive adversarial post-training for real-time interactive video generation.
\newblock In \emph{Advances in Neural Information Processing Systems}, 2025.

\bibitem[Lipman et~al.(2023)Lipman, Chen, Ben-Hamu, Nickel, and Le]{lipman2023flow}
Yaron Lipman, Ricky T.~Q. Chen, Heli Ben-Hamu, Maximilian Nickel, and Matt Le.
\newblock Flow matching for generative modeling.
\newblock In \emph{International Conference on Learning Representations}, 2023.

\bibitem[Liu et~al.(2025)Liu, Hu, Xu, Shan, and Lu]{liu2025rolling}
Kunhao Liu, Wenbo Hu, Jiale Xu, Ying Shan, and Shijian Lu.
\newblock Rolling forcing: Autoregressive long video diffusion in real time.
\newblock \emph{arXiv preprint arXiv:2509.25161}, 2025.

\bibitem[Lu and Song(2024)]{lu2024simplifying}
Cheng Lu and Yang Song.
\newblock Simplifying, stabilizing and scaling continuous-time consistency models.
\newblock \emph{arXiv preprint arXiv:2410.11081}, 2024.

\bibitem[Lu et~al.(2026)Lu, Zuo, and Deng]{lu2026raven}
Yanzuo Lu, Ronglai Zuo, and Jiankang Deng.
\newblock Raven: Real-time autoregressive video extrapolation with consistency-model grpo.
\newblock \emph{arXiv preprint arXiv:2605.15190}, 2026.

\bibitem[Peebles and Xie(2023)]{peebles2023scalable}
William Peebles and Saining Xie.
\newblock Scalable diffusion models with transformers.
\newblock In \emph{IEEE/CVF International Conference on Computer Vision}, 2023.

\bibitem[Polyak et~al.(2024)Polyak, Zohar, Brown, Tjandra, Sinha, Lee, Vyas, Shi, Ma, Chuang, et~al.]{polyak2024movie}
Adam Polyak, Amit Zohar, Andrew Brown, Andros Tjandra, Animesh Sinha, Ann Lee, Apoorv Vyas, Bowen Shi, Chih-Yao Ma, Ching-Yao Chuang, et~al.
\newblock {Movie Gen}: A cast of media foundation models.
\newblock \emph{arXiv preprint arXiv:2410.13720}, 2024.

\bibitem[Qian et~al.(2026)Qian, Wang, Zhang, Zou, Yu, Li, Liu, Li, Kang, Huang, An, Zhang, Jiang, Wang, Sun, Liu, and Li]{qian2026matrixgame35}
Runjia Qian, Zile Wang, Jihai Zhang, Kai Zou, Wei Yu, Jiaxing Li, Zexiang Liu, Yaokun Li, Fei Kang, Kaichen Huang, Mengyin An, Haobo Zhang, Biao Jiang, Jiahua Wang, Haofeng Sun, Yang Liu, and Yangguang Li.
\newblock {Matrix-Game 3.5}: Enhancing real-time streaming interactive world models with patch memory.
\newblock \emph{arXiv preprint arXiv:2608.29910}, 2026.

\bibitem[Ren et~al.(2025)Ren, Shen, Huang, Ling, Lu, Nimier-David, M{\"u}ller, Keller, Fidler, and Gao]{ren2025gen3c}
Xuanchi Ren, Tianchang Shen, Jiahui Huang, Huan Ling, Yifan Lu, Merlin Nimier-David, Thomas M{\"u}ller, Alexander Keller, Sanja Fidler, and Jun Gao.
\newblock {GEN3C}: 3d-informed world-consistent video generation with precise camera control.
\newblock In \emph{IEEE/CVF Conference on Computer Vision and Pattern Recognition}, 2025.

\bibitem[{Robbyant Team} et~al.(2026){Robbyant Team}, Gao, Wang, Zeng, Zhu, Cheng, Li, Wang, Xu, Ma, Chen, Liu, Cheng, Yao, Zhu, Meng, Zheng, Bai, Chen, Shen, Yu, Zhu, Shen, and Ouyang]{gao2026lingbot}
{Robbyant Team}, Zelin Gao, Qiuyu Wang, Yanhong Zeng, Jiapeng Zhu, Ka~Leong Cheng, Yixuan Li, Hanlin Wang, Yinghao Xu, Shuailei Ma, Yihang Chen, Jie Liu, Yansong Cheng, Yao Yao, Jiayi Zhu, Yihao Meng, Kecheng Zheng, Qingyan Bai, Jingye Chen, Zehong Shen, Yue Yu, Xing Zhu, Yujun Shen, and Hao Ouyang.
\newblock Advancing open-source world models.
\newblock \emph{arXiv preprint arXiv:2601.20540}, 2026.

\bibitem[Savva et~al.(2026)Savva, Michel, Lu, Waiwitlikhit, Meehan, Mishra, Poddar, Lu, and Xie]{savva2026solaris}
Georgy Savva, Oscar Michel, Daohan Lu, Suppakit Waiwitlikhit, Timothy Meehan, Dhairya Mishra, Srivats Poddar, Jack Lu, and Saining Xie.
\newblock Solaris: Building a multiplayer video world model in minecraft.
\newblock \emph{arXiv preprint arXiv:2602.22208}, 2026.

\bibitem[Singer et~al.(2022)Singer, Polyak, Hayes, Yin, An, Zhang, Hu, Yang, Ashual, Gafni, et~al.]{singer2022make}
Uriel Singer, Adam Polyak, Thomas Hayes, Xi Yin, Jie An, Songyang Zhang, Qiyuan Hu, Harry Yang, Oron Ashual, Oran Gafni, et~al.
\newblock Make-a-video: Text-to-video generation without text-video data.
\newblock \emph{arXiv preprint arXiv:2209.14792}, 2022.

\bibitem[Sohl-Dickstein et~al.(2015)Sohl-Dickstein, Weiss, Maheswaranathan, and Ganguli]{sohldickstein2015deep}
Jascha Sohl-Dickstein, Eric Weiss, Niru Maheswaranathan, and Surya Ganguli.
\newblock Deep unsupervised learning using nonequilibrium thermodynamics.
\newblock In \emph{International Conference on Machine Learning}, pages 2256--2265, 2015.

\bibitem[Song et~al.(2025)Song, Chen, Simchowitz, Du, Tedrake, and Sitzmann]{song2025history}
Kiwhan Song, Boyuan Chen, Max Simchowitz, Yilun Du, Russ Tedrake, and Vincent Sitzmann.
\newblock History-guided video diffusion.
\newblock \emph{arXiv preprint arXiv:2502.06764}, 2025.

\bibitem[Song et~al.(2021)Song, Sohl-Dickstein, Kingma, Kumar, Ermon, and Poole]{song2021score}
Yang Song, Jascha Sohl-Dickstein, Diederik~P. Kingma, Abhishek Kumar, Stefano Ermon, and Ben Poole.
\newblock Score-based generative modeling through stochastic differential equations.
\newblock In \emph{International Conference on Learning Representations}, 2021.

\bibitem[Song et~al.(2023)Song, Dhariwal, Chen, and Sutskever]{song2023consistency}
Yang Song, Prafulla Dhariwal, Mark Chen, and Ilya Sutskever.
\newblock Consistency models.
\newblock In \emph{International Conference on Machine Learning}, 2023.

\bibitem[Su et~al.(2026)Su, Zhuang, Li, Lu, Li, Li, Wu, Jin, Zhang, Huang, et~al.]{su2026sgf+}
Zihan Su, Junhao Zhuang, Yaowei Li, Siwen Lu, Haoran Li, Lingen Li, Haoyu Wu, Weiyang Jin, Songchun Zhang, Haoyang Huang, et~al.
\newblock Sgf+: Decoupling gradient flows for autoregressive video generation.
\newblock \emph{arXiv preprint arXiv:2610.10429}, 2026.

\bibitem[Sun et~al.(2025)Sun, Zhang, Wang, Wu, Wang, Wang, Wang, Zhang, Wang, and Guo]{sun2025worldplay}
Wenqiang Sun, Haiyu Zhang, Haoyuan Wang, Junta Wu, Zehan Wang, Zhenwei Wang, Yunhong Wang, Jun Zhang, Tengfei Wang, and Chunchao Guo.
\newblock {WorldPlay}: Towards long-term geometric consistency for real-time interactive world modeling.
\newblock \emph{arXiv preprint arXiv:2512.14614}, 2025.

\bibitem[{Team Wan} et~al.(2025){Team Wan}, Wang, Ai, Wen, Mao, Xie, Chen, Yu, Zhao, Yang, Zeng, Wang, Zhang, Zhou, Wang, Chen, Zhu, Zhao, Yan, Huang, Feng, Zhang, Li, Wu, Chu, Feng, Zhang, Sun, Fang, Wang, Gui, Weng, Shen, Lin, Wang, Wang, Zhou, Wang, Shen, Yu, Shi, Huang, Xu, Kou, Lv, Li, Liu, Wang, Zhang, Huang, Li, Wu, Liu, Pan, Zheng, Hong, Shi, Feng, Jiang, Han, Wu, and Liu]{wan2025}
{Team Wan}, Ang Wang, Baole Ai, Bin Wen, Chaojie Mao, Chen-Wei Xie, Di Chen, Feiwu Yu, Haiming Zhao, Jianxiao Yang, Jianyuan Zeng, Jiayu Wang, Jingfeng Zhang, Jingren Zhou, Jinkai Wang, Jixuan Chen, Kai Zhu, Kang Zhao, Keyu Yan, Lianghua Huang, Mengyang Feng, Ningyi Zhang, Pandeng Li, Pingyu Wu, Ruihang Chu, Ruili Feng, Shiwei Zhang, Siyang Sun, Tao Fang, Tianxing Wang, Tianyi Gui, Tingyu Weng, Tong Shen, Wei Lin, Wei Wang, Wei Wang, Wenmeng Zhou, Wente Wang, Wenting Shen, Wenyuan Yu, Xianzhong Shi, Xiaoming Huang, Xin Xu, Yan Kou, Yangyu Lv, Yifei Li, Yijing Liu, Yiming Wang, Yingya Zhang, Yitong Huang, Yong Li, You Wu, Yu Liu, Yulin Pan, Yun Zheng, Yuntao Hong, Yupeng Shi, Yutong Feng, Zeyinzi Jiang, Zhen Han, Zhi-Fan Wu, and Ziyu Liu.
\newblock {Wan}: Open and advanced large-scale video generative models.
\newblock \emph{arXiv preprint arXiv:2503.20314}, 2025.

\bibitem[Teng et~al.(2025)Teng, Jia, Sun, Li, Li, Tang, Han, Zhang, Zhang, Luo, et~al.]{teng2025magi}
Hansi Teng, Hongyu Jia, Lei Sun, Lingzhi Li, Maolin Li, Mingqiu Tang, Shuai Han, Tianning Zhang, WQ Zhang, Weifeng Luo, et~al.
\newblock {MAGI-1}: Autoregressive video generation at scale.
\newblock \emph{arXiv preprint arXiv:2505.13211}, 2025.

\bibitem[Valevski et~al.(2024)Valevski, Leviathan, Arar, and Fruchter]{valevski2024gamengen}
Dani Valevski, Yaniv Leviathan, Moab Arar, and Shlomi Fruchter.
\newblock Diffusion models are real-time game engines.
\newblock \emph{arXiv preprint arXiv:2408.14837}, 2024.

\bibitem[Wang et~al.(2023)Wang, Lu, Wang, Bao, Li, Su, and Zhu]{wang2023prolificdreamer}
Zhengyi Wang, Cheng Lu, Yikai Wang, Fan Bao, Chongxuan Li, Hang Su, and Jun Zhu.
\newblock {ProlificDreamer}: High-fidelity and diverse text-to-3d generation with variational score distillation.
\newblock In \emph{Advances in Neural Information Processing Systems}, 2023.

\bibitem[Wang et~al.(2024)Wang, Yuan, Wang, Li, Chen, Xia, Luo, and Shan]{wang2024motionctrl}
Zhouxia Wang, Ziyang Yuan, Xintao Wang, Yaowei Li, Tianshui Chen, Menghan Xia, Ping Luo, and Ying Shan.
\newblock {MotionCtrl}: A unified and flexible motion controller for video generation.
\newblock In \emph{ACM SIGGRAPH}, 2024.

\bibitem[Weissenborn et~al.(2019)Weissenborn, T{\"a}ckstr{\"o}m, and Uszkoreit]{weissenborn2019scaling}
Dirk Weissenborn, Oscar T{\"a}ckstr{\"o}m, and Jakob Uszkoreit.
\newblock Scaling autoregressive video models.
\newblock \emph{arXiv preprint arXiv:1906.02634}, 2019.

\bibitem[Williams and Zipser(1989)]{williams1989learning}
Ronald~J. Williams and David Zipser.
\newblock A learning algorithm for continually running fully recurrent neural networks.
\newblock \emph{Neural Computation}, 1\penalty0 (2):\penalty0 270--280, 1989.

\bibitem[Wu et~al.(2021)Wu, Huang, Zhang, Li, Ji, Yang, Sapiro, and Duan]{wu2021godiva}
Chenfei Wu, Lun Huang, Qianxi Zhang, Binyang Li, Lei Ji, Fan Yang, Guillermo Sapiro, and Nan Duan.
\newblock Godiva: Generating open-domain videos from natural descriptions.
\newblock \emph{arXiv preprint arXiv:2104.14806}, 2021.

\bibitem[Wu et~al.(2022)Wu, Liang, Ji, Yang, Fang, Jiang, and Duan]{wu2022nuwa}
Chenfei Wu, Jian Liang, Lei Ji, Fan Yang, Yuejian Fang, Daxin Jiang, and Nan Duan.
\newblock N{\"u}wa: Visual synthesis pre-training for neural visual world creation.
\newblock In \emph{European conference on computer vision}, pages 720--736. Springer, 2022.

\bibitem[Xiao et~al.(2025)Xiao, Zhang, Meng, Wang, Zhang, and Zhang]{xiao2025knot}
Steven Xiao, XIndi Zhang, Dechao Meng, Qi Wang, Peng Zhang, and Bang Zhang.
\newblock Knot forcing: Taming autoregressive video diffusion models for real-time infinite interactive portrait animation.
\newblock \emph{arXiv preprint arXiv:2512.21734}, 2025.

\bibitem[Xing et~al.(2024)Xing, Xia, Zhang, Chen, Yu, Liu, Liu, Wang, Shan, and Wong]{xing2024dynamicrafter}
Jinbo Xing, Menghan Xia, Yong Zhang, Haoxin Chen, Wangbo Yu, Hanyuan Liu, Gongye Liu, Xintao Wang, Ying Shan, and Tien-Tsin Wong.
\newblock Dynamicrafter: Animating open-domain images with video diffusion priors.
\newblock In \emph{European Conference on Computer Vision}, pages 399--417. Springer, 2024.

\bibitem[Yan et~al.(2021)Yan, Zhang, Abbeel, and Srinivas]{yan2021videogpt}
Wilson Yan, Yunzhi Zhang, Pieter Abbeel, and Aravind Srinivas.
\newblock Videogpt: Video generation using vq-vae and transformers.
\newblock \emph{arXiv preprint arXiv:2104.10157}, 2021.

\bibitem[Yang et~al.(2025)Yang, Huang, Chu, Xiao, Zhao, Wang, Li, Xie, Chen, Lu, et~al.]{yang2025longlive}
Shuai Yang, Wei Huang, Ruihang Chu, Yicheng Xiao, Yuyang Zhao, Xianbang Wang, Muyang Li, Enze Xie, Yingcong Chen, Yao Lu, et~al.
\newblock {LongLive}: Real-time interactive long video generation.
\newblock \emph{arXiv preprint arXiv:2509.22622}, 2025.

\bibitem[Yang et~al.(2024)Yang, Teng, Zheng, Ding, Huang, Xu, Yang, Hong, Zhang, Feng, et~al.]{yang2024cogvideox}
Zhuoyi Yang, Jiayan Teng, Wendi Zheng, Ming Ding, Shiyu Huang, Jiazheng Xu, Yuanming Yang, Wenyi Hong, Xiaohan Zhang, Guanyu Feng, et~al.
\newblock {CogVideoX}: Text-to-video diffusion models with an expert transformer.
\newblock \emph{arXiv preprint arXiv:2408.06072}, 2024.

\bibitem[Yesiltepe et~al.(2026)Yesiltepe, Meral, Akan, Oktay, and Yanardag]{yesiltepe2026infinity}
Hidir Yesiltepe, Tuna Meral, Adil~Kaan Akan, Kaan Oktay, and Pinar Yanardag.
\newblock Infinity-rope: Action-controllable infinite video generation emerges from autoregressive self-rollout.
\newblock In \emph{Proceedings of the IEEE/CVF Conference on Computer Vision and Pattern Recognition}, pages 40256--40265, 2026.

\bibitem[Yi et~al.(2025)Yi, Jang, Cho, Nam, Yoon, and Kim]{yi2025deep}
Jung Yi, Wooseok Jang, Paul~Hyunbin Cho, Jisu Nam, Heeji Yoon, and Seungryong Kim.
\newblock Deep forcing: Training-free long video generation with deep sink and participative compression.
\newblock \emph{arXiv preprint arXiv:2512.05081}, 2025.

\bibitem[Yin et~al.(2024{\natexlab{a}})Yin, Gharbi, Park, Zhang, Shechtman, Durand, and Freeman]{yin2024improved}
Tianwei Yin, Micha{\"e}l Gharbi, Taesung Park, Richard Zhang, Eli Shechtman, Fredo Durand, and William~T. Freeman.
\newblock Improved distribution matching distillation for fast image synthesis.
\newblock In \emph{Advances in Neural Information Processing Systems}, 2024{\natexlab{a}}.

\bibitem[Yin et~al.(2024{\natexlab{b}})Yin, Gharbi, Zhang, Shechtman, Durand, Freeman, and Park]{yin2024one}
Tianwei Yin, Micha{\"e}l Gharbi, Richard Zhang, Eli Shechtman, Fredo Durand, William~T. Freeman, and Taesung Park.
\newblock One-step diffusion with distribution matching distillation.
\newblock In \emph{IEEE/CVF Conference on Computer Vision and Pattern Recognition}, 2024{\natexlab{b}}.

\bibitem[Yin et~al.(2025)Yin, Zhang, Zhang, Freeman, Durand, Shechtman, and Huang]{yin2025causvid}
Tianwei Yin, Qiang Zhang, Richard Zhang, William~T. Freeman, Fr{\'e}do Durand, Eli Shechtman, and Xun Huang.
\newblock From slow bidirectional to fast autoregressive video diffusion models.
\newblock In \emph{IEEE/CVF Conference on Computer Vision and Pattern Recognition}, 2025.

\bibitem[Ying et~al.(2026)Ying, Hu, Ren, Li, Chen, Wang, Cao, Cai, and Ding]{ying2026wbench}
Kaining Ying, Hengrui Hu, Siyu Ren, Jiamu Li, Fengjiao Chen, Ziwen Wang, Xuezhi Cao, Xunliang Cai, and Henghui Ding.
\newblock {WBench}: A comprehensive multi-turn benchmark for interactive video world model evaluation.
\newblock \emph{arXiv preprint arXiv:2605.25874}, 2026.

\bibitem[Zhang et~al.(2026{\natexlab{a}})Zhang, Sun, Wang, Wu, Zhang, Wang, Qiao, and Guo]{zhang2026worldplay2}
Haiyu Zhang, Wenqiang Sun, Tengfei Wang, Junta Wu, Jun Zhang, Yunhong Wang, Yu Qiao, and Chunchao Guo.
\newblock {WorldPlay2}: Extending real-time interactive world models in control and horizon.
\newblock \emph{arXiv preprint arXiv:2609.35560}, 2026{\natexlab{a}}.

\bibitem[Zhang et~al.(2026{\natexlab{b}})Zhang, Jiang, Chen, Wang, Liu, Li, Chen, Lin, Zhou, Jin, et~al.]{zhang2026vidu2}
Jintao Zhang, Kai Jiang, Jintao Chen, Xu Wang, Deyuan Liu, Jungang Li, Dechuang Chen, Ming Lin, Jingjiang Zhou, Haopeng Jin, et~al.
\newblock Vidu s2: Real-time interactive, editable, and spatial video generation.
\newblock \emph{arXiv preprint arXiv:2609.11638}, 2026{\natexlab{b}}.

\bibitem[Zhang et~al.(2026{\natexlab{c}})Zhang, Jiang, Chen, Wang, Luo, Wang, Chen, Li, Ye, Chen, et~al.]{zhang2026vidu}
Jintao Zhang, Kai Jiang, Jintao Chen, Xu Wang, Yang Luo, Yuji Wang, Dechuang Chen, Jungang Li, Chengyang Ye, Marco Chen, et~al.
\newblock Vidu s1: A real-time interactive video generation model.
\newblock \emph{arXiv preprint arXiv:2607.03118}, 2026{\natexlab{c}}.

\bibitem[Zhang et~al.(2026{\natexlab{d}})Zhang, Qian, Kong, Jiang, Wang, Chen, Li, Chen, Wu, Chen, et~al.]{zhang2026matrix}
Jihai Zhang, Runjia Qian, Kaiqin Kong, Biao Jiang, Puyi Wang, Siru Chen, Tianle Li, Guanjie Chen, Junyi Wu, Chubin Chen, et~al.
\newblock Matrix-game 4.0: Harnessing video models for real-time playable game worlds.
\newblock \emph{arXiv preprint arXiv:2610.27999}, 2026{\natexlab{d}}.

\bibitem[Zhang et~al.(2025)Zhang, Peng, Wang, Wang, Zhu, Kang, Jiang, Gao, Li, Liu, and Zhou]{zhang2025matrixgame}
Yifan Zhang, Chunli Peng, Boyang Wang, Puyi Wang, Qingcheng Zhu, Fei Kang, Biao Jiang, Zedong Gao, Eric Li, Yang Liu, and Yahui Zhou.
\newblock {Matrix-Game}: Interactive world foundation model.
\newblock \emph{arXiv preprint arXiv:2506.18701}, 2025.

\bibitem[Zhao et~al.(2024)Zhao, Zhu, Xiang, Zheng, Li, and Zhu]{zhao2024identifying}
Min Zhao, Hongzhou Zhu, Chendong Xiang, Kaiwen Zheng, Chongxuan Li, and Jun Zhu.
\newblock Identifying and solving conditional image leakage in image-to-video diffusion model.
\newblock \emph{Advances in Neural Information Processing Systems}, 37:\penalty0 30300--30326, 2024.

\bibitem[Zhao et~al.(2025{\natexlab{a}})Zhao, He, Chen, Zhu, Li, and Zhu]{zhao2025riflex}
Min Zhao, Guande He, Yixiao Chen, Hongzhou Zhu, Chongxuan Li, and Jun Zhu.
\newblock Riflex: A free lunch for length extrapolation in video diffusion transformers.
\newblock \emph{arXiv preprint arXiv:2502.15894}, 2025{\natexlab{a}}.

\bibitem[Zhao et~al.(2025{\natexlab{b}})Zhao, Zhu, Wang, Yan, Zhang, He, Yang, Li, and Zhu]{zhao2025ultravico}
Min Zhao, Hongzhou Zhu, Yingze Wang, Bokai Yan, Jintao Zhang, Guande He, Ling Yang, Chongxuan Li, and Jun Zhu.
\newblock Ultravico: Breaking extrapolation limits in video diffusion transformers.
\newblock \emph{arXiv preprint arXiv:2511.20123}, 2025{\natexlab{b}}.

\bibitem[Zhao et~al.(2026{\natexlab{a}})Zhao, Zhu, Yan, Zhou, Chen, Wang, Sun, Fang, Xun, Li, Zheng, He, Yang, Li, Bao, and Zhu]{zhao2026minwm}
Min Zhao, Hongzhou Zhu, Bokai Yan, Zihan Zhou, Yimin Chen, Honglie Wang, Wenqiang Sun, Zhengwei Fang, Zizheng Xun, Zihao Li, Kaiwen Zheng, Guande He, Xiao Yang, Chongxuan Li, Fan Bao, and Jun Zhu.
\newblock {minWM}: A full-stack open-source framework for real-time interactive video world models.
\newblock \emph{arXiv preprint arXiv:2605.30263}, 2026{\natexlab{a}}.

\bibitem[Zhao et~al.(2026{\natexlab{b}})Zhao, Zhu, Zheng, Zhou, Yan, Li, Yang, Li, and Zhu]{zhao2026causalforcingpp}
Min Zhao, Hongzhou Zhu, Kaiwen Zheng, Zihan Zhou, Bokai Yan, Xinyuan Li, Xiao Yang, Chongxuan Li, and Jun Zhu.
\newblock Causal forcing++: Scalable few-step autoregressive diffusion distillation for real-time interactive video generation.
\newblock \emph{arXiv preprint arXiv:2605.15141}, 2026{\natexlab{b}}.

\bibitem[Zheng et~al.(2026)Zheng, He, Zhao, Zhang, Chen, Chen, Lin, Liu, Zhu, and Ma]{zheng2026causalrcm}
Kaiwen Zheng, Guande He, Min Zhao, Jintao Zhang, Huayu Chen, Jianfei Chen, Chen-Hsuan Lin, Ming-Yu Liu, Jun Zhu, and Qianli Ma.
\newblock Causal-{rCM}: A unified teacher-forcing and self-forcing open recipe for autoregressive diffusion distillation in streaming video generation and interactive world models.
\newblock \emph{arXiv preprint arXiv:2606.25473}, 2026.

\bibitem[Zheng et~al.(2024)Zheng, Peng, Yang, Shen, Li, Liu, Zhou, Li, and You]{zheng2024open}
Zangwei Zheng, Xiangyu Peng, Tianji Yang, Chenhui Shen, Shenggui Li, Hongxin Liu, Yukun Zhou, Tianyi Li, and Yang You.
\newblock Open-sora: Democratizing efficient video production for all.
\newblock \emph{arXiv preprint arXiv:2412.20404}, 2024.

\bibitem[Zhou and Miao(2026)]{zhou2026astronex}
Xin Zhou and Cong Miao.
\newblock Astronex-world 1.0: Real-time interactive world model foundation.
\newblock \emph{arXiv preprint arXiv:2609.20034}, 2026.

\bibitem[Zhu et~al.(2026{\natexlab{a}})Zhu, Liu, Zhao, Ye, Chen, Yu, He, Han, and Xie]{zhu2026sanawm}
Haoyi Zhu, Haozhe Liu, Yuyang Zhao, Tian Ye, Junsong Chen, Jincheng Yu, Tong He, Song Han, and Enze Xie.
\newblock {SANA-WM}: Efficient minute-scale world modeling with hybrid linear diffusion transformer.
\newblock \emph{arXiv preprint arXiv:2605.15178}, 2026{\natexlab{a}}.

\bibitem[Zhu et~al.(2026{\natexlab{b}})Zhu, Zhao, He, Su, Li, and Zhu]{zhu2026causalforcing}
Hongzhou Zhu, Min Zhao, Guande He, Hang Su, Chongxuan Li, and Jun Zhu.
\newblock Causal forcing: Autoregressive diffusion distillation done right for high-quality real-time interactive video generation.
\newblock \emph{arXiv preprint arXiv:2602.02214}, 2026{\natexlab{b}}.

\bibitem[Zhuang et~al.(2026)Zhuang, Zhang, Bian, Li, Luo, Liu, Jin, Zhang, He, Zhang, Li, Huang, Xue, and Duan]{zhuang2026self}
Junhao Zhuang, Shiyi Zhang, Yuxuan Bian, Yaowei Li, Yawen Luo, Yijun Liu, Weiyang Jin, Songchun Zhang, Xianglong He, Xuying Zhang, Haoran Li, Haoyang Huang, Zeyue Xue, and Nan Duan.
\newblock Self gradient forcing: Native long video extrapolation.
\newblock \emph{arXiv preprint arXiv:2607.20368}, 2026.

\end{thebibliography}
}
\clearpage
\appendix
\maketitlesupplementary

\section{Extended Related Work}
\label{app:related_work}

This section reviews related work on video generation, autoregressive video
diffusion and its few-step distillation, and their applications to
interactive video models.

\paragraph{Video generation.}
Recent advances in video
generation~\cite{lin2024open,zheng2024open,yang2024cogvideox,bao2024vidu,zhao2024identifying,zhao2025riflex,zhao2025ultravico}
mainly follow two paradigms: diffusion
models~\cite{blattmann2023align,he2022latent,ho2022imagen,singer2022make,chen2023videocrafter1,xing2024dynamicrafter,hacohen2024ltx,polyak2024movie},
such as HunyuanVideo~\cite{kong2024hunyuanvideo}, Wan~\cite{wan2025}, and
Vidu~\cite{bao2024vidu}, and autoregressive
models~\cite{wu2021godiva,hong2022cogvideo,wu2022nuwa,yan2021videogpt,weissenborn2019scaling},
such as VideoGPT~\cite{yan2021videogpt} and
VideoPoet~\cite{kondratyuk2023videopoet}.
Bidirectional diffusion models achieve high visual fidelity, but generate all
frames jointly, leading to high first-frame latency and limited
interactivity.
Discrete autoregressive models instead generate incrementally and thus support
real-time interaction, but generally lag behind in visual quality.
To combine the strengths of both, autoregressive video diffusion
models~\cite{jin2024pyramidal,teng2025magi,chen2025skyreels,deng2024autoregressive},
such as MAGI-1~\cite{teng2025magi} and SkyReels-V2~\cite{chen2025skyreels},
generate a video block by block and denoise each block with diffusion.

\paragraph{Few-step autoregressive distillation.}
To further reduce latency, recent works distill autoregressive diffusion
models into few-step
generators~\cite{yin2025causvid,huang2025selfforcing,zhu2026causalforcing,zhao2026causalforcingpp,zheng2026causalrcm,zhuang2026self,lin2026autoregressive,bandyopadhyay2026cmd,savva2026solaris},
such as CausVid~\cite{yin2025causvid}, Self
Forcing~\cite{huang2025selfforcing}, Causal
Forcing~\cite{zhu2026causalforcing}, Solaris~\cite{savva2026solaris},
RAVEN~\cite{lu2026raven}, Self Gradient Forcing~\cite{zhuang2026self}, and
SGF+~\cite{su2026sgf+}, enabling real-time generation.
Building on these methods, many works further extend autoregressive video
diffusion to long-form
generation~\cite{yesiltepe2026infinity,yi2025deep,yang2025longlive,cui2025self,liu2025rolling,infinite-forcing,cui2026lol,huang2026live}.

\paragraph{Causal teachers in autoregressive distillation.}
Mainstream distillation
methods~\cite{yin2025causvid,huang2025selfforcing,zhu2026causalforcing,zhao2026causalforcingpp}
supervise the causal student with a bidirectional teacher via DMD.
Attempts to use a causal teacher instead have been less successful:
CausVid~\cite{yin2025causvid} and Causal
Forcing++~\cite{zhao2026causalforcingpp} find causal teachers inferior to bidirectional
ones, but both train the student on ground-truth context rather than on its
own rollouts, which by itself induces exposure bias.
The closest work to ours is CMD~\cite{bandyopadhyay2026cmd}, which reports
gains of a causal teacher over a bidirectional one; however, its causal
teacher is further fine-tuned from a pretrained bidirectional model, without a
reported matched budget.
Moreover, CMD argues that each block should be scored under the context in
which the student generated it, and thus adopts scoring against a separate,
slightly noised copy of the rollout, Full CMD (\cref{app:tf_style_dmd}), as
its main method, whereas we find scoring under high-noise context to be more
effective (\cref{sec:rethink}).

\paragraph{Interactive video models.}
Autoregressive video diffusion and its distillation have substantially
advanced real-time interactive video models, including camera-controllable
world models such as Genie 3~\cite{genie3}, WorldPlay~\cite{sun2025worldplay},
minWM~\cite{zhao2026minwm}, DreamX-World 1.0~\cite{team2026dreamx},
Matrix-Game 3.5~\cite{qian2026matrixgame35}, Astronex-World
1.0~\cite{zhou2026astronex}, Zing-0.5~\cite{chen2026zing},
ZYT-World~\cite{hu2026zyt}, and Matrix-Game 4.0~\cite{zhang2026matrix}, as
well as real-time interactive avatar systems such as Knot
Forcing~\cite{xiao2025knot}, Live Avatar~\cite{huang2025live}, and
Vidu-S~\cite{zhang2026vidu,zhang2026vidu2}.

\section{Method Details}
\label{app:method}

\subsection{Supervision Conditions of \method and SF}
\label{app:appendix_bi_dmd}

This section details the \emph{causal supervision} property of \method
discussed in \cref{sec:method}: we show that \method supervises each block
using only the past blocks and the controls up to that block, whereas
SF~\cite{huang2025selfforcing} additionally conditions on future blocks and
controls.
\method updates the causal student through per-block score differences on its
on-policy rollout $\hat{\vx}^{1:N}$ (\cref{eq:df_dmd_gradient}), while SF
scores the whole rollout with a bidirectional teacher and critic.
We therefore rewrite the SF gradient in the same per-block form and compare
what the supervision of each block conditions on.

\paragraph{Causal Self Forcing (\method).}
Since the teacher and critic of \method are block-causal, their outputs for
block $i$ in a single forward pass depend only on the inputs of blocks up to
$i$.
The score difference of block $i$ (\cref{eq:df_dmd_score}) thus conditions
only on the past blocks and the controls up to block $i$:
\begin{equation}
    \begin{aligned}
        \Delta s_i^{\text{\tiny CSF}}
         & = s_{\mathrm{fake}}(\hat{\vx}_{t_i}^{i},t_i\mid\hat{\vx}_{\vt}^{<i},\vt^{<i},\vc^{\leq i})    \\
         & \quad - s_{\mathrm{real}}(\hat{\vx}_{t_i}^{i},t_i\mid
        \underbrace{\hat{\vx}_{\vt}^{<i},\vt^{<i},\vc^{\leq i}}_{\text{past}}),
    \end{aligned}
    \label{eq:app_csf_score}
\end{equation}
which involves no future blocks or controls.

\paragraph{Self Forcing (SF).}
SF perturbs the rollout with a shared timestep $t$ into
$\hat{\vx}_t^{1:N}=(1-t)\hat{\vx}^{1:N}+t\vepsilon^{1:N}$ and scores it
jointly with the bidirectional teacher and critic.
By the chain rule, its gradient in \cref{eq:dmd_gradient} decomposes over
blocks as
\begin{equation}
    \nabla_\theta\mathcal{L}_{\mathrm{DMD}}
    =\mathbb{E}\!\left[\sum_{i=1}^{N}w(t)
        \left(\frac{\partial\hat{\vx}^i}{\partial\theta}\right)^{\!\top}
        \Delta s_i^{\text{\tiny SF}}\right],
    \label{eq:app_bi_dmd_decomp}
\end{equation}
where $\Delta s_i^{\text{\tiny SF}}$ is the block-$i$ component of the joint score
difference
$s_{\mathrm{fake}}(\hat{\vx}_t^{1:N},t,\vc^{1:N})-s_{\mathrm{real}}(\hat{\vx}_t^{1:N},t,\vc^{1:N})$.
To express this component as a block-level score, we take the teacher score
$s_{\mathrm{real}}$ as an example; the derivation for the critic score
$s_{\mathrm{fake}}$ is identical.
Isolating block $i$ from the remaining blocks
$\hat{\vx}_t^{\neq i}=(\hat{\vx}_t^{<i},\hat{\vx}_t^{>i})$, we factorize the
perturbed teacher density as
\begin{equation}
    \begin{aligned}
        p_{\mathrm{real},t}(\hat{\vx}_t^{1:N}\mid\vc^{1:N})
         & = p_{\mathrm{real},t}(\hat{\vx}_t^i\mid\hat{\vx}_t^{\neq i},\vc^{1:N}) \\
         & \quad\cdot p_{\mathrm{real},t}(\hat{\vx}_t^{\neq i}\mid\vc^{1:N}).
    \end{aligned}
    \label{eq:app_cond_factorization}
\end{equation}
Taking the logarithm and differentiating with respect to $\hat{\vx}_t^i$, the
second factor drops out since it does not depend on $\hat{\vx}_t^i$:
\begin{equation}
    \begin{aligned}
        \big[s_{\mathrm{real}}(\hat{\vx}_t^{1:N},t,\vc^{1:N})\big]_i
         & = \nabla_{\hat{\vx}_t^i}\log p_{\mathrm{real},t}(\hat{\vx}_t^{1:N}\mid\vc^{1:N})        \\
         & = \nabla_{\hat{\vx}_t^i}\log p_{\mathrm{real},t}(\hat{\vx}_t^i\mid\hat{\vx}_t^{\neq i},\vc^{1:N}),
    \end{aligned}
    \label{eq:app_block_score_identity}
\end{equation}
\ie, the block-$i$ component of the joint score is the conditional score of
block $i$ given all other blocks, including the future ones.
Applying the same derivation to the critic density $p_{\mathrm{fake},t}$
yields the per-block score difference
\begin{equation}
    \begin{aligned}
        \Delta s_i^{\text{\tiny SF}}
         & = s_{\mathrm{fake}}(\hat{\vx}_{t}^{i},t\mid\hat{\vx}_{t}^{<i},\vc^{\leq i},
        \textcolor{cred}{\hat{\vx}_{t}^{>i},\vc^{>i}})                  \\
         & \quad - s_{\mathrm{real}}(\hat{\vx}_{t}^{i},t\mid
        \underbrace{\hat{\vx}_{t}^{<i},\vc^{\leq i}}_{\text{past}},
        \textcolor{cred}{\underbrace{\hat{\vx}_{t}^{>i},\vc^{>i}}_{\text{future}}}).
    \end{aligned}
    \label{eq:bi_dmd_score}
\end{equation}

\paragraph{Comparison.}
When generating block $i$, the causal student
$G_\theta(\cdot\mid\hat{\vx}^{<i},\vc^{\leq i})$ observes only the past blocks
and the controls up to block $i$.
Under SF, however, the supervision of this block varies with the future blocks
$\hat{\vx}_t^{>i}$ and controls $\vc^{>i}$ (red in \cref{eq:bi_dmd_score}),
which the student cannot observe.
\method instead supervises each block with exactly the information available
to the student (\cref{eq:app_csf_score}), consistent with the autoregressive
factorization in \cref{eq:ar_factorization}, and requires no future frames or
controls.

\subsection{Details of CMD}
\label{app:tf_style_dmd}

CMD~\cite{bandyopadhyay2026cmd} fine-tunes a pretrained bidirectional model
into a multi-step causal teacher with the DF objective, initializes the
few-step student from this teacher, and applies DMD on Self-Forcing
rollouts~\cite{huang2025selfforcing}.
To score a rollout $\hat{\vx}^{1:N}$ with the teacher and critic, CMD starts
from a base variant (Base CMD) and then moves to a refined one (Full CMD),
which matches the scoring context to the context under which the student
generated each block and is adopted as its main method:

\emph{Base CMD} perturbs the rollout into a single noisy sequence, in which
each noisy target also serves as the context of later targets.

\emph{Full CMD} instead scores the noisy targets against a separate copy of
the rollout, perturbed with a small fixed noise level, as context.
Concretely, each target block is perturbed with its own timestep $t_i$ as
$\hat{\vx}^i_{t_i}=(1-t_i)\hat{\vx}^i+t_i\vepsilon^i$, while a second copy of
the rollout is perturbed with a single fixed context noise level
$\rho\in[0,1)$ as $\hat{\vx}^i_{\rho}=(1-\rho)\hat{\vx}^i+\rho\vepsilon'^i$.
Under a block-causal mask, the target of block $i$ attends only to the context
blocks before it, which gives the per-block score difference
\begin{equation}
    \begin{aligned}
        \Delta s_i^{\text{\tiny TF}}
         & = s_{\mathrm{fake}}(\hat{\vx}_{t_i}^{i},t_i\mid\hat{\vx}_{\rho}^{<i},\rho\vone,\vc^{\leq i})        \\
         & \quad - s_{\mathrm{real}}(\hat{\vx}_{t_i}^{i},t_i\mid\hat{\vx}_{\rho}^{<i},\rho\vone,\vc^{\leq i}).
    \end{aligned}
    \label{eq:tf_dmd_score}
\end{equation}
CMD sets $\rho$ to scheduler timestep 256, \ie, $\rho=0.256$.
Full CMD is thus an instance of the TF-style DMD in \cref{sec:rethink}: the
teacher and critic process $2N$ blocks ($2\times$ tokens), and $\rho=0$
reduces it to clean-context TF-style DMD.
Notably, CMD's own ablation~\cite{bandyopadhyay2026cmd} is consistent with
our analysis in \cref{sec:rethink}: scoring against the clean rollout
($\rho=0$) lowers the VBench-I2V Total score from $88.33$ (Base CMD) to
$87.25$.

In this paper, clean-context TF-style DMD is applied to both the TF and DF
teachers in \cref{fig:sec3_wbench}, Full CMD serves as a baseline in
\cref{tab:main_world_models}, and \cref{tab:ablation_tf_dmd} varies $\rho$
with our DF teacher.

\subsection{Algorithm of \method}
\label{app:algorithm}

We take the SGF~\cite{zhuang2026self} setting used in our experiments as an
example; the pseudocode of \method training is given in \cref{alg:csf}.

\begin{algorithm*}[t]
    \caption{Causal Self Forcing (\method) training with SGF}
    \label{alg:csf}
    \small
    \begin{algorithmic}[1]
        \Require Frozen causal DF teacher $s_{\mathrm{real}}$; initialized
        causal student $G_\theta$ and DF critic $s_{\mathrm{fake}}$; $T$
        student denoising timesteps $\tau_1>\cdots>\tau_T$
        \While{not converged}
        \Statex \textit{// No-gradient self-rollout}
        \State Sample exit step $k\sim\mathcal{U}\{1,\ldots,T\}$
        \Comment{random exit}
        \For{$i=1,\ldots,N$}
        \Comment{no gradient}
        \State $\vz_{\tau_1}^{i}\sim\mathcal{N}(\vzero,\mI)$
        \For{$j=1,\ldots,k-1$}
        \State $\vz_{\tau_{j+1}}^{i}\gets(1-\tau_{j+1})\,
            G_\theta(\vz_{\tau_j}^{i},\tau_j\mid\hat{\vx}^{<i},\vc^{\leq i})
            +\tau_{j+1}\vepsilon'$,\; $\vepsilon'\sim\mathcal{N}(\vzero,\mI)$
        \Comment{denoise and re-noise}
        \EndFor
        \State $\hat{\vx}^{i}\gets G_\theta(\vz_{\tau_k}^{i},\tau_k\mid
            \hat{\vx}^{<i},\vc^{\leq i})$
        \Comment{exit at $\tau_k$}
        \State Write clean $\hat{\vx}^{i}$ into the KV cache
        \EndFor
        \Statex \textit{// Generator/critic update}
        \State Sample $\vt$ and $\vepsilon^{1:N}\sim\mathcal{N}(\vzero,\mI)$
        \If{generator step}
        \State $\hat{\vx}^{1:N}\gets G_\theta\big(\vz_{\tau_k}^{1:N},\tau_k\mid
            \mathrm{sg}(\hat{\vx}^{1:N}),\vc^{1:N}\big)$
        \Comment{parallel context-gradient reconstruction}
        \State $\hat{\vx}_{t_i}^{i}\gets(1-t_i)\hat{\vx}^{i}+t_i\vepsilon^{i}$,
        $i=1,\ldots,N$
        \State $\Delta s_i\gets
            s_{\mathrm{fake}}(\hat{\vx}_{t_i}^{i},t_i\mid
            \hat{\vx}_{\vt}^{<i},\vt^{<i},\vc^{\leq i})
            -s_{\mathrm{real}}(\hat{\vx}_{t_i}^{i},t_i\mid
            \hat{\vx}_{\vt}^{<i},\vt^{<i},\vc^{\leq i})$, $i=1,\ldots,N$
        \Comment{one block-causal pass}
        \State Update $G_\theta$ with
        $\nabla_\theta\mathcal{L}=\sum_{i=1}^{N}w(t_i)
            \big(\partial\hat{\vx}^{i}/\partial\theta\big)^{\!\top}\Delta s_i$
        \Else
        \State $\hat{\vx}_{t_i}^{i}\gets(1-t_i)\,\mathrm{sg}(\hat{\vx}^{i})
            +t_i\vepsilon^{i}$, $i=1,\ldots,N$
        \State Update $s_{\mathrm{fake}}$ with
        $\nabla_{\theta_{\mathrm{fake}}}\sum_{i=1}^{N}\lambda(t_i)\big\|v_{\mathrm{fake}}(\hat{\vx}_{t_i}^{i},t_i\mid
            \hat{\vx}_{\vt}^{<i},\vt^{<i},\vc^{\leq i})
            -\big(\vepsilon^{i}-\mathrm{sg}(\hat{\vx}^{i})\big)\big\|_2^2$
        \Comment{$v_{\mathrm{fake}}$: velocity form of $s_{\mathrm{fake}}$}
        \EndIf
        \EndWhile
    \end{algorithmic}
\end{algorithm*}

\subsection{AnyFlow Initialization}
\label{app:anyflow}

SolarWM-5B~\cite{huang2026solarwm} converts its bidirectional teacher into a
causal student initialization with AnyFlow~\cite{gu2026anyflow}.
Whereas our DF Stage~1 learns the instantaneous velocity under randomly
corrupted context, AnyFlow trains a TF model ($2\times$ tokens) to predict a
flow map, \ie, the average velocity $u_\theta(\vx_t,t,r)$ from $t$ to any
$r\leq t$, such that $\vx_r=\vx_t-(t-r)\,u_\theta(\vx_t,t,r)$.
Following MeanFlow, it is supervised through the identity
$u_\theta(\vx_t,t,r)+(t-r)\frac{d}{dt}u_\theta(\vx_t,t,r)=\vv(\vx_t,t)$,
where $\vv$ is the CFG-guided velocity and the total derivative along the
trajectory is estimated by finite differences.
For $r=t$, the derivative term vanishes and the objective reduces to flow
matching; SolarWM-5B uses $r=t$ for half of the batch and $r<t$ for the rest.

\paragraph{Training cost.}
We measure training compute in units of a DF forward pass, counting a backward
pass as two forward passes; a bidirectional forward costs one unit, and a TF
forward costs two units due to its $2\times$ tokens.
A DF or bidirectional training step thus costs 3 units, and a TF training step
6 units.
Besides its TF training forward, each AnyFlow sample with $r<t$ requires four
no-gradient TF forwards (2 units each): one for the unconditional prediction in
CFG, one for the instantaneous velocity $u_\theta(\vx_t,t,t)$, and two for the
finite-difference derivative.
For $r=t$, the velocity coincides with the training forward and the derivative
is not needed, leaving only the unconditional forward.
The two cases cost $6+4\times2=14$ and $6+2=8$ units, \ie, $11$ units per
sample on average, or about $3.7\times$ a DF training step.

\section{Experimental Details}
\label{app:appendix_experiments}

\paragraph{Hyperparameters.}
\Cref{tab:hyperparameters} lists the hyperparameters of both stages, which
are identical to those of the corresponding stages of
SolarWM-5B~\cite{huang2026solarwm} (its bidirectional-teacher stage for
Stage~1 and its distillation stage for Stage~2), except for the Stage-2
global batch size.
For efficiency, all distillation runs, including baselines and ablations, use
a global batch size of 64 instead of 256, which performs similarly in our
experiments.
All models are trained on H800 GPUs with BF16 mixed precision and FSDP.

\begin{table}[t]
    \centering
    \small
    \caption{\textbf{Training hyperparameters of the two stages of \method.}}
    \label{tab:hyperparameters}
    \setlength{\tabcolsep}{4pt}
    \begin{tabular}{lcc}
        \toprule
                                 & Stage~1            & Stage~2                               \\
                                 & (DF training)      & (Distillation)                        \\
        \midrule
        Optimizer                & AdamW              & AdamW                                 \\
        $(\beta_1,\beta_2)$      & $(0.9,0.95)$       & $(0,0.999)$                           \\
        Weight decay             & $0.01$             & $0$                                   \\
        (Student) Learning rate            & $5\times10^{-5}$   & $2\times10^{-6}$                      \\
        Critic learning rate     & --                 & $4\times10^{-7}$                      \\
        Warmup steps             & 1k                 & --                                    \\
        Global batch size        & 256                & 64                                    \\
        Gradient clipping        & $1.0$              & $10$                                  \\
        EMA decay                & $0.9999$           & $0.99$                                \\
        Timestep shift $s$       & 5                  & 5                                     \\
        Critic updates per step  & --                 & 5                                     \\
        Student timesteps        & --                 & $\{1000,750,500,250\}$                \\
        Teacher CFG scale        & --                 & 3                                     \\
        Distill.\ timestep range & --                 & $[0.02,0.98]$                         \\
        \bottomrule
    \end{tabular}
\end{table}

\begin{table*}[t]
    \centering
    \small
    \setlength{\tabcolsep}{3pt}
    \caption{\textbf{Per-dimension VBench~\cite{huang2024vbench} scores on the
            SANA-WM benchmark~\cite{zhu2026sanawm}.} SC, BC, TF, MS, AQ, IQ,
        DD, and OC denote subject consistency, background consistency,
        temporal flickering, motion smoothness, aesthetic quality, imaging
        quality, dynamic degree, and overall consistency; Overall is the
        VBench quality score aggregated over all dimensions except OC,
        reported as VBench in \cref{tab:sana_wm}.
        \textbf{Bold}: best; \underline{underline}: second best.}
    \label{tab:sana_wm_vbench}
    \begin{tabular*}{\textwidth}{@{\extracolsep{\fill}}lccccccccc@{}}
        \toprule
        Model & SC$\uparrow$ & BC$\uparrow$ & TF$\uparrow$ & MS$\uparrow$ & AQ$\uparrow$
        & IQ$\uparrow$ & DD$\uparrow$ & OC$\uparrow$ & Overall$\uparrow$ \\
        \midrule
        \multicolumn{10}{@{}l}{\emph{Simple-trajectory split}} \\
        HY-WorldPlay~\cite{sun2025worldplay} & \textbf{87.55} & \textbf{91.44} & \textbf{98.00} & \textbf{98.97} & \underline{56.71} & \underline{69.13} & 23.75 & \underline{24.40} & 77.46 \\
        LingBot-World-Fast~\cite{gao2026lingbot} & 82.52 & \underline{89.13} & 95.21 & 96.97 & 54.79 & 65.80 & \textbf{92.50} & \textbf{24.83} & \underline{78.34} \\
        \multicolumn{10}{@{}l}{Base Model: Wan2.2-5B} \\
        \quad+~CausVid~\cite{yin2025causvid} & 68.53 & 82.36 & 93.45 & 93.54 & 41.90 & 63.20 & 85.00 & 21.41 & 68.91 \\
        \quad+~SF (SolarWM-5B~\cite{huang2026solarwm}) & 74.98 & 85.42 & 96.04 & 97.34 & 49.22 & 55.97 & 87.50 & 23.63 & 73.99 \\
        \quad+~Full CMD~\cite{bandyopadhyay2026cmd} & 76.74 & 84.86 & 95.81 & 97.62 & 52.98 & 69.01 & \underline{88.75} & 22.67 & 76.93 \\
        \quad+~\method (ours) & \underline{82.89} & 88.37 & \underline{96.60} & \underline{98.10} & \textbf{57.22} & \textbf{70.54} & 65.00 & 23.66 & \textbf{78.40} \\
        \midrule
        \multicolumn{10}{@{}l}{\emph{Hard-trajectory split}} \\
        HY-WorldPlay~\cite{sun2025worldplay} & \textbf{87.29} & \textbf{91.03} & \textbf{97.26} & \textbf{98.87} & \textbf{55.97} & \underline{69.69} & 46.25 & \underline{24.38} & \underline{78.67} \\
        LingBot-World-Fast~\cite{gao2026lingbot} & \underline{81.90} & \underline{88.41} & 95.05 & 96.94 & 54.20 & 66.43 & \underline{97.50} & \textbf{24.73} & 78.39 \\
        \multicolumn{10}{@{}l}{Base Model: Wan2.2-5B} \\
        \quad+~CausVid~\cite{yin2025causvid} & 70.12 & 83.11 & 93.65 & 93.41 & 43.65 & 66.80 & 85.00 & 21.98 & 70.18 \\
        \quad+~SF (SolarWM-5B~\cite{huang2026solarwm}) & 75.57 & 85.98 & 95.53 & 97.19 & 50.33 & 57.30 & \underline{97.50} & 23.96 & 75.07 \\
        \quad+~Full CMD~\cite{bandyopadhyay2026cmd} & 76.52 & 85.56 & 95.04 & 97.38 & 52.44 & 68.96 & \textbf{98.75} & 22.98 & 77.26 \\
        \quad+~\method (ours) & 81.35 & 87.65 & \underline{95.65} & \underline{97.86} & \underline{55.79} & \textbf{70.84} & 92.50 & 23.82 & \textbf{79.40} \\
        \bottomrule
    \end{tabular*}

    \vspace{4mm}
    \par
    \footnotesize
    \setlength{\tabcolsep}{1.5pt}
    \begin{minipage}[t]{\columnwidth}
    \centering
    \caption{\textbf{WBench navigation scores of TF teachers.}
        Teacher-forcing (TF) teachers at different pretraining steps are
        sampled in multiple steps with CFG$=3$.
        Metric abbreviations follow \cref{tab:ablation_rollout}.}
    \label{tab:ablation_tf_teacher_sampling}
    \begin{tabular*}{\columnwidth}{@{\extracolsep{\fill}}lcccccc@{}}
        \toprule
        Teacher steps & Overall$\uparrow$ & Qua.$\uparrow$ & Con.$\uparrow$ & Int.$\uparrow$ & Set.$\uparrow$ & Phy.$\uparrow$ \\
        \midrule
        20k & 69.4 & 73.9 & 77.3 & \textbf{77.1} & 76.0 & 42.4 \\
        30k & \textbf{72.8} & 75.1 & 79.0 & \textbf{77.1} & \textbf{76.1} & \textbf{56.6} \\
        40k & 71.9 & 75.0 & \textbf{80.3} & 75.6 & 74.6 & 54.0 \\
        50k & 71.1 & \textbf{75.3} & 80.0 & 73.5 & 73.6 & 53.0 \\
        \bottomrule
    \end{tabular*}
    \end{minipage}\hfill
    \begin{minipage}[t]{\columnwidth}
    \centering
    \caption{\textbf{WBench navigation scores of students distilled from TF
        teachers.}
        Students are distilled with clean-context TF-style DMD ($\rho=0$) from
        the TF teachers in \cref{tab:ablation_tf_teacher_sampling}.
        Metric abbreviations follow \cref{tab:ablation_rollout}.}
    \label{tab:ablation_tf_teacher_steps}
    \begin{tabular*}{\columnwidth}{@{\extracolsep{\fill}}lcccccc@{}}
        \toprule
        Teacher steps & Overall$\uparrow$ & Qua.$\uparrow$ & Con.$\uparrow$ & Int.$\uparrow$ & Set.$\uparrow$ & Phy.$\uparrow$ \\
        \midrule
        20k & 70.3 & 74.6 & 77.5 & 80.7 & 75.0 & 43.7 \\
        30k & \textbf{73.9} & \textbf{76.7} & 80.9 & \textbf{81.9} & 75.3 & \textbf{54.5} \\
        40k & 73.5 & 76.5 & 80.7 & 80.7 & \textbf{75.5} & 54.1 \\
        50k & 72.6 & 76.2 & \textbf{81.1} & 80.5 & 74.5 & 50.7 \\
        \bottomrule
    \end{tabular*}
    \end{minipage}
\end{table*}

\paragraph{Training budget.}
Measured in DF training steps (\cref{app:anyflow}), a bidirectional step costs
about the same, a TF step $2\times$, and an AnyFlow~\cite{gu2026anyflow} step
about $3.7\times$.
The 30k bidirectional and 3k AnyFlow steps of SolarWM-5B thus amount to about
41k DF-step equivalents, comparable to the 40k steps of our Stage~1.
Likewise, the TF teacher in \cref{sec:rethink} is trained for 20k steps to
match the compute of the 40k-step DF teacher. 
All distillation runs, including those of the baselines, are trained for up
to 1k steps until convergence, a cost small compared with teacher training.

\paragraph{Baselines.}
CausVid uses the same bidirectional teacher and AnyFlow initialization as
SolarWM-5B, while Full CMD initializes the teacher, critic, and student from
our Stage-1 checkpoint and applies TF-style DMD (\cref{app:tf_style_dmd}) with
$\rho=0.256$.
For HY-WorldPlay~\cite{sun2025worldplay}, we use its released 8B
autoregressive distilled model; for LingBot-World-Fast~\cite{gao2026lingbot},
we use its first-generation 14B model with local attention size 18 and
attention sink size 6.

\paragraph{Inference.}
All Wan2.2-5B-based models follow the inference setting of
SolarWM-5B~\cite{huang2026solarwm} on both WBench and the SANA-WM benchmark.
Each block is sampled with the four student timesteps in
\cref{tab:hyperparameters} using the EMA weights and a fixed noise seed.
For videos longer than the 21-latent-frame training clips, the KV cache keeps
only the most recent 18 latent frames without attention sinks, and the
temporal RoPE positions stop advancing at the end of the 21-frame training
horizon, while the camera PRoPE still follows the actual frame index.
Multi-step teachers are sampled with the UniPC solver in 50 steps.

\paragraph{WBench.}
We follow the official evaluation code of WBench~\cite{ying2026wbench} on its
navigation split (158 multi-turn cases with official 6-DoF camera
trajectories), and report the unweighted mean of the five dimensions as the
Overall score.
VLM-judged metrics use the self-hosted Qwen3.8-27B judge, so the scores are
not directly comparable to those obtained with the default Doubao judge.

\paragraph{SANA-WM benchmark.}
We follow the official evaluation protocol of the SANA-WM
benchmark~\cite{zhu2026sanawm}; \cref{tab:sana_wm_vbench} reports the
per-dimension VBench scores.

\section{Additional Analysis of Causal Teachers}
\label{app:teacher_analysis}

This section provides further evidence for the analysis in
\cref{sec:rethink}, covering distillation from TF teachers, the multi-step
sampling of DF teacher checkpoints, the context noise at which a DF teacher
is queried, and the sampling of DMD timesteps.

\paragraph{Teacher-forcing teachers.}
We first study distillation from TF teachers, which are trained only with
clean context, using clean-context TF-style DMD
(\cref{tab:ablation_tf_teacher_sampling,tab:ablation_tf_teacher_steps}).
First, the student stays close to its teacher at every checkpoint, exceeding
it by at most $1.6$ points; this small gain possibly stems from the
mode-seeking behavior of the reverse KL objective.
Second, under matched training compute, the TF teacher yields a much weaker
student: the 20k-step TF teacher, which matches the compute of the 40k-step
DF teacher since TF training uses $2\times$ tokens and is the one used in
\cref{fig:sec3_wbench}, gives a student of $70.3$, far below \method
($75.7$); even with more training compute, the best TF student ($73.9$ at
30k steps) still falls short of \method.

\paragraph{Diffusion-forcing teachers.}
\Cref{tab:ablation_df_teacher_steps} reports the multi-step sampling of the
DF teacher checkpoints in \cref{tab:ablation_teacher_steps}.
The teacher follows the same trend as its students, peaking at 40k steps.
Unlike the TF case, however, the \method student surpasses its DF teacher at
every checkpoint.
Moreover, under matched compute, the TF and DF teachers sample almost equally
well ($69.4$ vs.\ $69.6$), yet their students differ by $5.4$ points.
The gain of \method thus comes not from a stronger teacher, but from querying
the DF teacher under noisy context (\cref{sec:rethink}).

\paragraph{Context noise in distillation.}
We next study how the context noise at which a DF teacher is queried affects
distillation.
When distilling our DF teacher with TF-style DMD
(\cref{tab:ablation_tf_dmd}), the student generally improves with the
context noise $\rho$, from $68.2$ at $\rho=0$ to $74.5$ at $\rho=0.9$, while
Full CMD ($\rho=0.256$) reaches only $72.3$.
This confirms that the DF teacher is more reliable under high-noise context
(\cref{sec:rethink}), contrary to the choice of CMD.
\method outperforms all these variants ($75.7$) without tuning $\rho$ and
with half the scoring tokens.

\paragraph{Timestep sampling.}
For the DMD timesteps $\vt$ that perturb the rollout, we find that sharing
one timestep across blocks outperforms sampling independent per-block
timesteps ($75.7$ vs.\ $74.7$ in \cref{tab:ablation_timestep}), while both
remain above Full CMD.
We therefore use the shared timestep for all other results, whereas DF
pretraining keeps independent per-block timesteps.

\begin{table}[t]
    \centering
    \footnotesize
    \setlength{\tabcolsep}{1.5pt}
    \caption{\textbf{WBench navigation scores of DF teachers.}
        Diffusion-forcing (DF) teachers at different pretraining steps are
        sampled in multiple steps with CFG$=3$; their \method students are
        in \cref{tab:ablation_teacher_steps}.
        Metric abbreviations follow \cref{tab:ablation_rollout}.}
    \label{tab:ablation_df_teacher_steps}
    \begin{tabular*}{\columnwidth}{@{\extracolsep{\fill}}lcccccc@{}}
        \toprule
        Teacher steps & Overall$\uparrow$ & Qua.$\uparrow$ & Con.$\uparrow$ & Int.$\uparrow$ & Set.$\uparrow$ & Phy.$\uparrow$ \\
        \midrule
        10k & 56.1 & 60.0 & 66.5 & 66.4 & 59.7 & 27.9 \\
        20k & 65.3 & 71.1 & 72.8 & 74.8 & 75.6 & 32.0 \\
        30k & 68.5 & 74.3 & 77.6 & \textbf{76.0} & 75.7 & 39.0 \\
        40k & \textbf{69.6} & \textbf{74.5} & \textbf{78.0} & 73.9 & \textbf{77.6} & \textbf{44.1} \\
        50k & 69.0 & 74.1 & 77.8 & 74.5 & 75.0 & 43.4 \\
        60k & 68.4 & 72.9 & 77.7 & 74.4 & 73.8 & 43.0 \\
        \bottomrule
    \end{tabular*}
\end{table}

\begin{table}[t]
    \centering
    \footnotesize
    \setlength{\tabcolsep}{1.5pt}
    \caption{\textbf{WBench navigation scores under different context noise
        levels in distillation.}
        Students are distilled from our DF teacher with TF-style DMD
        (CMD~\cite{bandyopadhyay2026cmd}) under context noise levels
        $\rho\in[0,1)$; $\rho=0$ is clean-context TF-style DMD and
        $\rho=0.256$ is Full CMD.
        Metric abbreviations follow \cref{tab:ablation_rollout}.}
    \label{tab:ablation_tf_dmd}
    \begin{tabular*}{\columnwidth}{@{\extracolsep{\fill}}lcccccc@{}}
        \toprule
        Context noise $\rho$ & Overall$\uparrow$ & Qua.$\uparrow$ & Con.$\uparrow$ & Int.$\uparrow$ & Set.$\uparrow$ & Phy.$\uparrow$ \\
        \midrule
        0 & 68.2 & 75.2 & 73.6 & 79.1 & 74.7 & 38.2 \\
        0.128 & 71.8 & 76.7 & 78.9 & 79.9 & 76.9 & 46.5 \\
        0.256 (Full CMD) & 72.3 & 76.6 & 78.1 & 80.0 & 76.6 & 50.1 \\
        0.512 & 74.1 & 77.8 & 81.0 & 80.8 & \textbf{78.0} & 53.0 \\
        0.6 & 72.8 & 77.9 & 80.1 & 76.5 & 77.2 & 52.3 \\
        0.7 & 73.8 & 78.1 & 80.5 & 78.7 & 77.6 & 53.9 \\
        0.8 & 73.9 & 78.4 & \textbf{81.4} & 78.2 & 76.1 & \textbf{55.2} \\
        0.9 & \textbf{74.5} & \textbf{78.5} & 81.2 & \textbf{81.4} & 77.3 & 54.0 \\
        0.95 & 74.1 & 78.3 & 80.3 & 80.8 & 77.6 & 53.4 \\
        \bottomrule
    \end{tabular*}
\end{table}

\begin{table}[t]
    \centering
    \footnotesize
    \setlength{\tabcolsep}{1.5pt}
    \caption{\textbf{WBench navigation scores of \method with different DMD
        timestep sampling.}
        The DMD timesteps $\vt$, which perturb the student rollout, are
        shared across blocks or sampled independently per block; DF
        pretraining is unchanged.
        Metric abbreviations follow \cref{tab:ablation_rollout}.}
    \label{tab:ablation_timestep}
    \begin{tabular*}{\columnwidth}{@{\extracolsep{\fill}}lcccccc@{}}
        \toprule
        DMD timesteps $\vt$ & Overall$\uparrow$ & Qua.$\uparrow$ & Con.$\uparrow$ & Int.$\uparrow$ & Set.$\uparrow$ & Phy.$\uparrow$ \\
        \midrule
        Independent per block & 74.7 & 77.8 & 80.5 & 81.1 & \textbf{77.2} & 56.9 \\
        Shared across blocks & \textbf{75.7} & \textbf{78.5} & \textbf{81.7} & \textbf{82.8} & 77.1 & \textbf{58.6} \\
        \bottomrule
    \end{tabular*}
\end{table}

\begin{figure*}[t]
    \centering
    \includegraphics[width=\textwidth]{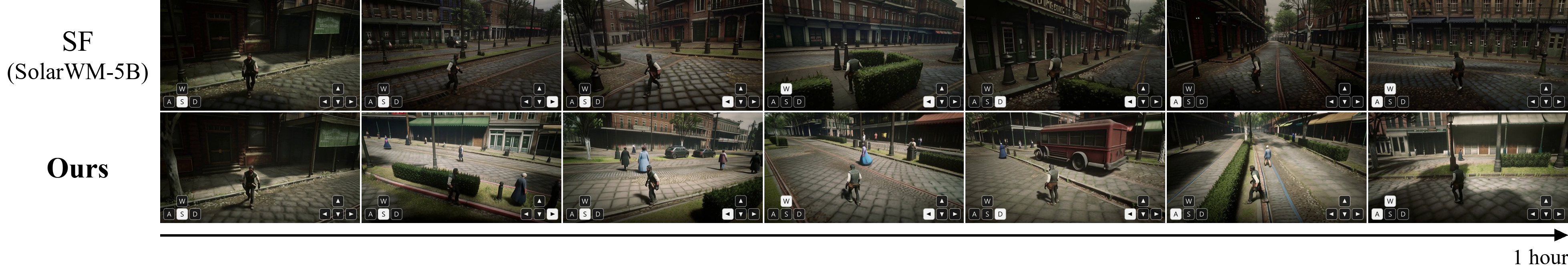}
    \caption{\textbf{One-hour rollout.} Rows correspond to methods and
        columns to time; both methods share the same input image and camera
        trajectory.
        \textbf{Prompt:} ``A cobblestone street runs through a historic
        district lined with multi-story brick buildings featuring wooden
        balconies, awnings, and large paneled doors. A grassy median with
        parallel metal tracks, trimmed hedges, and mature trees divides the
        roadway. Storefronts display faded painted signs and green-trimmed
        windows, with one entrance marked by vertical lettering. The pavement
        shows scattered leaves and debris. Overcast daylight casts soft
        shadows through the tree canopy, creating a muted, atmospheric
        ambiance throughout the corridor.''}
    \label{fig:supp_hour}
\end{figure*}

\section{More Qualitative Results}
\label{app:qualitative}

\Cref{fig:supp_hour} extends the rollout to one hour, $720\times$ the 5\,s
training length, and compares \method with the bidirectional-teacher baseline
SF.
Neither collapses, but SF becomes dark and low in contrast, and the character
gradually blurs, whereas \method keeps the character and the scene sharp
throughout the hour.

\begin{figure*}[t]
    \centering
    \includegraphics[width=\textwidth]{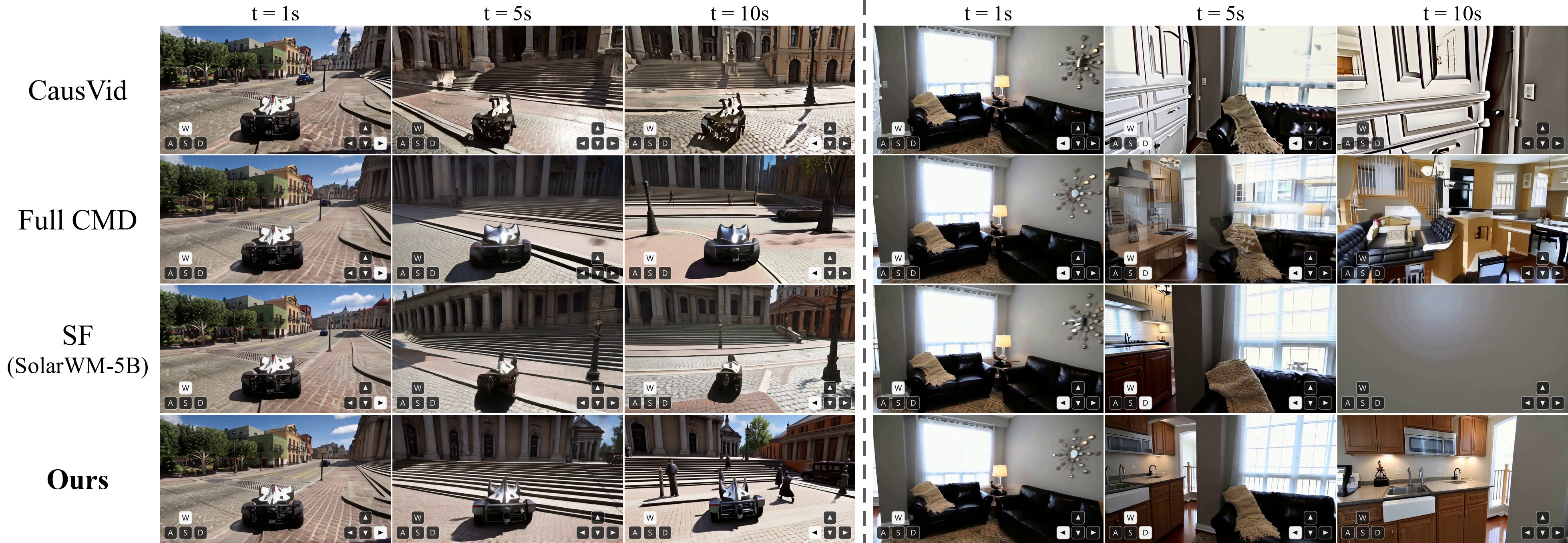}
    \caption{\textbf{Additional 10\,s rollouts.} Rows correspond to methods
        and columns to time; all methods share the same input image and
        camera trajectory.
        \textbf{Left prompt:} ``A sunlit cobblestone plaza stretches between
        colorful colonial buildings with stucco facades in green, yellow,
        orange, and cream. A grand neoclassical building with tall fluted
        columns and wide stone steps dominates one side, flanked by sculpted
        lion statues. Adjacent stands a historic stone church with red domes
        and arched bell towers. Trimmed trees line walkways, and ornate black
        lampposts punctuate the brick-paved sidewalks. The sky is bright blue
        with scattered white clouds, casting sharp shadows across the uneven
        stone surfaces.''
        \textbf{Right prompt:} ``A residential interior features a living
        area with two black leather sofas arranged on a patterned area rug,
        one draped with a cream throw and the other with a beige blanket. A
        lit table lamp with a white shade sits on a small dark table between
        the sofas, beside a decorative bird figurine. Sheer white curtains
        cover a large window. Gray walls display a circular metallic wall
        sculpture with mirror elements. Beyond a partial wall, an open dining
        area contains a dark round table with four metal chairs, a wooden
        sideboard, and a chandelier with multiple lights. Two arched windows
        and a staircase with white railings are visible. Dark hardwood
        flooring extends throughout. A kitchen counter with a sink and light
        cabinets appears at the edge of the dining space.''}
    \label{fig:supp_short}
\end{figure*}

\begin{figure*}[t]
    \centering
    \includegraphics[width=\textwidth]{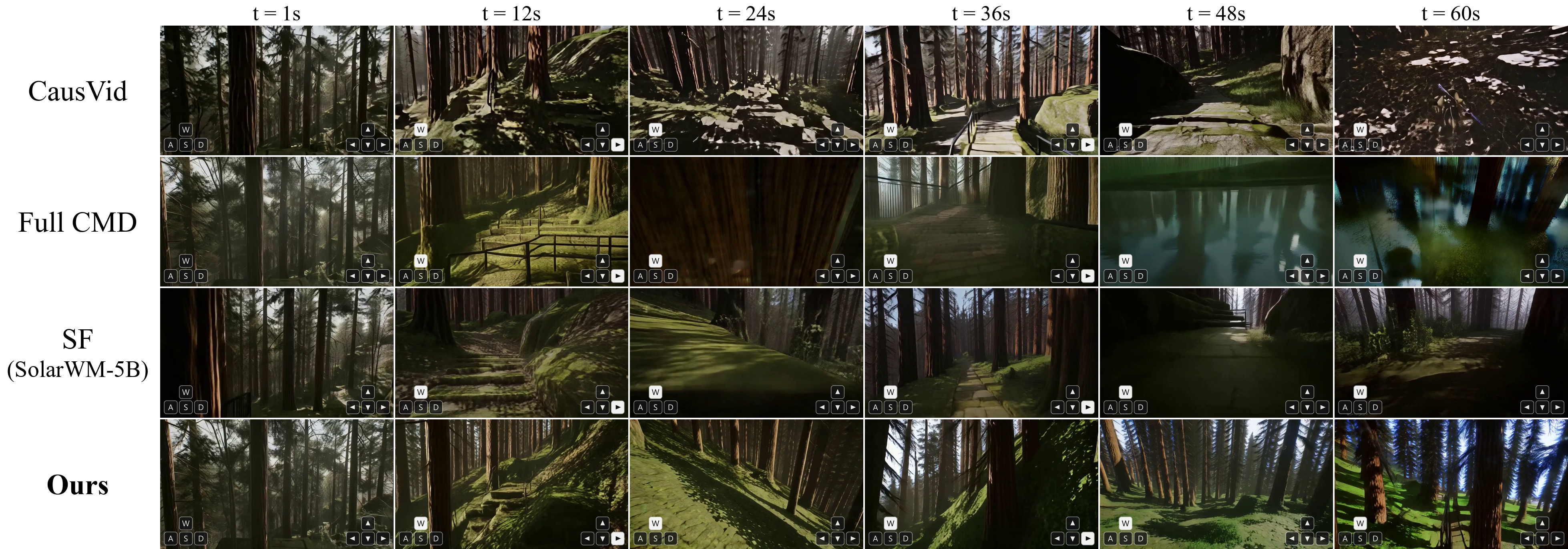}
    \caption{\textbf{Additional 60\,s rollout.} Rows correspond to methods
        and columns to time; all methods share the same input image and
        camera trajectory.
        \textbf{Prompt:} ``A dense coniferous forest with towering
        straight-trunked trees rises on both sides of a moss-covered stone
        pathway. The weathered flagstones and occasional stone steps form an
        elevated trail that winds through rocky terrain, with patches of
        green moss and low vegetation growing between cracks. Large mossy
        boulders line the path, and simple black metal railings appear at
        steeper sections. Sunlight filters through the canopy, casting
        dappled shadows across the stones and creating soft volumetric light
        in the air. The atmosphere is misty and serene, with muted greens,
        browns, and grays dominating the palette. The forest floor below the
        path is uneven, covered with grass, ferns, and scattered rocks.''}
    \label{fig:supp_long}
\end{figure*}

\Cref{fig:supp_short,fig:supp_long} provide additional comparisons with
CausVid, Full CMD, and SF (SolarWM-5B) on user-study cases not shown in
\cref{fig:qualitative}; all methods share the same input image and camera
trajectory.
In the 10\,s rollouts (\cref{fig:supp_short}), CausVid deforms the car in the
plaza, and SF collapses into a nearly uniform frame in the living room at
10\,s.
In the 60\,s forest rollout (\cref{fig:supp_long}), CausVid degrades into
high-contrast noise, Full CMD drifts into a blurry, water-like scene, and SF
becomes darker and loses detail.
\method preserves the scene layout and details in all cases, although its
colors become more saturated toward the end of the forest rollout.

\end{document}